\documentclass{ieeeaccess}
\usepackage{cite}
\usepackage{amsmath,amssymb,amsfonts}
\usepackage{algorithm} 
\usepackage{algpseudocode}
\usepackage{graphicx}
\usepackage{textcomp}

\usepackage{booktabs, makecell}
\usepackage{siunitx}
\usepackage{multirow}
\usepackage{multicol}
\usepackage{placeins}
\usepackage{colortbl}
\usepackage{xcolor}
\usepackage{float}
\usepackage{tabularx}
\usepackage{hyperref}
\usepackage{lscape}

\def\BibTeX{{\rm B\kern-.05em{\sc i\kern-.025em b}\kern-.08em
    T\kern-.1667em\lower.7ex\hbox{E}\kern-.125emX}}
\begin{document}
\history{Received 16 July 2022, accepted 2 August 2022, date of publication 5 August 2022, (this is a pre-print version of the published article).}
\doi{10.1109/ACCESS.2022.3196905}

\title{Modified Genetic Algorithm for Feature Selection and Hyper Parameter Optimization: Case of XGBoost in Spam Prediction}
\author{\uppercase{Nazeeh Ghatasheh}\authorrefmark{1},
\uppercase{Ismail Altaharwa\authorrefmark{2}, and Khaled Aldebei}.\authorrefmark{1}}
\address[1]{Department of Information Technology, The University of Jordan, Aqaba, Jordan (e-mail: \{n.ghatasheh, k.debei\}@ju.edu.jo)}
\address[2]{Department of Computer Information Systems, The University of Jordan, Aqaba, Jordan (e-mail: i\_taharwa@ju.edu.jo)}


\markboth
{Ghatasheh \headeretal: Modified GA for Feature Selection and Hyper Parameter Optimization in Spam Prediction}
{Ghatasheh \headeretal: Modified GA for Feature Selection and Hyper Parameter Optimization in Spam Prediction}

\corresp{Corresponding author: Nazeeh Ghatasheh (e-mail: n.ghatasheh@ju.edu.jo).}

\begin{abstract}
Recently, spam on online social networks has attracted attention in the research and business world. Twitter has become the preferred medium to spread spam content. Many research efforts attempted to encounter social networks spam. Twitter brought extra challenges represented by the feature space size, and imbalanced data distributions. Usually, the related research works focus on part of these main challenges or produce black-box models. In this paper, we propose a modified genetic algorithm for simultaneous dimensionality reduction and hyper parameter optimization over imbalanced datasets. The algorithm initialized an eXtreme Gradient Boosting classifier and reduced the features space of tweets dataset; to generate a spam prediction model. The model is validated using a 50 times repeated 10-fold stratified cross-validation, and analyzed using nonparametric statistical tests. The resulted prediction model attains on average 82.32\% and 92.67\% in terms of geometric mean and accuracy respectively, utilizing less than 10\% of the total feature space. The empirical results show that the modified genetic algorithm outperforms $Chi^2$ and $PCA$ feature selection methods. In addition, eXtreme Gradient Boosting outperforms many machine learning algorithms, including BERT-based deep learning model, in spam prediction. Furthermore, the proposed approach is applied to SMS spam modeling and compared to related works.

\end{abstract}

\begin{keywords}
Genetic Algorithm, Business Analytics, eXtreme Gradient Boosting, Feature Selection, Hyper Parameter Optimization, Spam Prediction.

\end{keywords}

\titlepgskip=-15pt

\maketitle

\section{Introduction}
\label{sec:introduction}

\PARstart{S}{pam} remains one of the long lasting security threats. E-mail spams represent a true challenge against mail service providers at the early stages of the Internet. Web spams exploit social engineering to lure a privileged user to login into a deceptive service. As Internet users developed awareness skills and became more competent to distinguish fake web content from truly legitimate one, attackers exploit the pervasiveness of social networks and corresponding media to launch the latest generation of spams, namely social spam. In addition to the opportunity to target a larger number of victims, social networks create an environment for ever-evolving avenues for spammers. It goes beyond traditional individual compromising activities such as monetary frauds towards large-scale campaigns. Quite recently released Twitter dataset distinguished more than five ways of twitter spams, including, but not limited to, profanity, insulting, hate speech, malicious links, fraudulent reviews \cite{Misc:utkmi_TwitterCompitition19}. Similarly, recent research efforts considered similar spamming approaches against other online social networks and short message service (SMS) \cite{article:Liu21,Article:Chowdhury20}. It is not surprising that twitter reviews spam policy periodically~\cite{Misc:TwitterSpamPolicy20}.
\par
As social spam campaigns emerged as a contemporary challenge against users, companies and even more governments, countermeasures evolved in a hand raising contest fashion. Earlier solutions were limited to the rule-based and regular expression matching. However, as spammers developed good experience to evade such detectors, information security practitioners considered content-based characteristics restrictively. Contemporary mature solutions utilize both content-based and account-based characteristics. In most cases, the ultimate goal is to find the shortest list of characteristics or features that indicate spamming behavior~\cite{Article:Wu18,Proc:Benjamin09}. Some studies step further to identify spammers themselves~\cite{Article:Mukunthan21,Proc:Gao12}. Machine learning techniques are leveraged in many ways to develop detection models. Earlier models utilized straightforward classification and categorization algorithms such as Support Vector MAchine (SVM), Naïve Bayes (NB), K-Nearest Neighbor (K-NN), and Decision Trees (DT)~\cite{Article:Chowdhury20,Article:Wu18,Proc:Benjamin09}. More advanced solutions explore opportunities of improvement as a result of utilizing deep learning (DL) techniques~\cite{article:Liu21,Article:ALOM20}.
\par
Deep learning based solutions approved to outperform conventional machine learning based prediction models. Considering social spams, such behavior of deep learning models is justifiable as it performs well in identifying local patterns~\cite{Article:ALOM20}. However, such performance comes at the expenses of model complexity. Additionally, Artificial Neural Networks (ANNs) models of deep learning are hard to interpret. Scalability and interpretability remain two contradictory desired characteristics of any social spam detector. In order to tackle this issue, we propose a novel dimension reduction solution. As parameter tuning is an unavoidable task regardless of the nature of the underlying prediction model, the proposed solution leverages a genetic algorithm to tune the parameters of the prediction model and select best descriptive features simultaneously. Such generated prediction models are still interpretable utilizing the final set of retained features. Further, proposed architecture allows developers to choose among a wide range of granularity depending on their targets and underlying computation capabilities. 
\par
A wide range of optimization techniques are proposed in literature. Alatas and Bingol categorized intelligent optimization techniques according to their scientific basis~\cite{alatas2019physics}. Further they compared their performance to the light-based intelligent optimizers~\cite{alatas2020comparative}. Genetic algorithms (GA), biology based optimizer, is the most popular type of evolutionary algorithms (EA) for parameter optimization. It demonstrates noticeably outstanding performance for a wide range of problems. Genetic algorithms retain merits of both metaheuristic search algorithms and stochastic optimization techniques~\cite{akyol2017plant}. This combination enables genetic algorithms to reach a global optima within a relatively fewer number of generations compared to other evolutionary algorithms.
\par
One of the major issues in spam text research is the limited availability of labeled text datasets with high quality \cite{kaddoura2022systematic,borse2022state}. For example the well known benchmark datasets are few, and many researchers use tools to collect domain specific datasets. As a result, many of the available text datasets have limited number of attributes, unverified class labeling, related to a specific language, imbalanced class distributions, or biased data. Furthermore, currently social media facilitate sharing multimedia content (e.g.,audio, video, text, images, etc) but incorporating such mix of content in model building seems to be one of the future challenges. Table \ref{tab:literature} shows a summary of the related research works in Tweets modeling and points to some research gaps.

\begin{table*}[]
\centering
\caption{Summary of related research works}
\label{tab:literature}
\begin{tabular}{p{0.03\linewidth} p{0.22\linewidth} p{0.2\linewidth} p{0.22\linewidth} p{0.22\linewidth}}
\toprule
Ref.&
  Work Done &
  Classifiers &
  Claims &
  Notes \\
\midrule
\cite{benitez2018improved} &
  Improved Genetic Algorithm for feature selection &
  Multinomial Naive Bayes &
  Outperforming Information Gain and CHI square for feature selection &
  Maximum reduction of features is more than 50\%. Reports only total accuracy. No information regarding model validation \\
\cite{utama2019sentiment} &
  Mutual information method for feature selection, Predefined number of features is tested &
  Naïve Bayes, SVM, Regression Logistic, Decision Tree &
  Naive Bayes classification was the fastest. SVM Linear Classifier attained highest accuracy &
  Manual setup of selected features size. No significant improvement over ANNOVA F and CHI square for feature selection. Train and Test dataset splits. Report Accuracy, SD, Error \% \\
\cite{prusa2015impact} &
  Used ten filter-based feature selection techniques, ANOVA tests &
  k-nearest neighbors, Logistic Regression, C4.5 decision tree, Multilayer Perceptron, ANN &
  200 is the optimal number of features to select &
  Reports AUC. 5-fold cross-validation \\
\cite{jain2019sentiment} &
  CFS Subset Evaluation and Information Gain feature selection &
  SVM, K-Nearest Neighbor, Naive Bayes, AdaBoost, Bagging algorithms &
  Accuracy 92.96\% by SVM &
  Focus on one domain, no basis for generalization. \\
\cite{wang2020optimal} &
  Trend-line analysis of the relationships between number of features and accuracy &
  Naïve Bayes, Maximum Entropy, SVM, Extreme Learning Machine &
  The relationship between the number features and the accuracy can be determined and it is independent of the machine learning approach &
  Used balanced datastes. Validated using percentage split (75\% training and 25\% testing) \\
\cite{saif2012alleviating} &
  Selected 42 to 34,855 features that represent a public tweets dataset &
  Naïve Bayes, Maximum Entropy &
  86.3\% Accuracy by ``Sentiment-topic features''. Selecting features beyond 500 will not improve the performance significantly &
  Report Accuracy only. No information about model validation \\
\cite{suchetha2019comparing} &
  Evolutionary computation based feature selection. Subset of 500, 1000, and 1500 features examined &
  Feature Selection: Information Gain feature, Particle Swarm, Ant Colony, Cuckoo Search, Firefly Search. Classification: LibLinear, K-Nearest Neighbor, Naive Bayes &
  LibLinear attained best performance. 78.6 AUC using 500 features by Firefly Search &
  5-fold cross-validation. Arbitrary selected features number. Reports AUC only \\
 \cite{jain2019optimizing} &
  Compare models and Feature selection &
  SVM, Naïve Bayes, ANN, K-Nearest Neighbor, Random Forest. Deep Learning: Recurrent NN, LSTM &
  Tweets Models: Accuracy 95.09 using LSTM. 5000 Features Accuracy 93.81 &
  Ham is the positive class. No information about model validation. Arbitrary feature selection. Minimum number of features selected 5000 \\
\cite{taneja2022comparison} &
  Assess transfer learning in predicting disaster tweets,  SMS spam, News Groups, and IMDB binary sentiment classification &
  DistilBERT, BERT, H2O AutoML, and 7 others &
  Accuracy 0.84\%, 0.98\%, 0.98\% in predicting disaster tweets, SMS spam, and IMDB respectively. &
  Reports Accuracy only. Percentage split validation (80\% training and 20\% validation). \\
\bottomrule
\end{tabular}
\end{table*}


\par
In order to evaluate the proposed solution, a real-world Twitter dataset is utilized, a quite large number of experiments incurred. Most experiments ended up with satisfactory performance due to the incorporation of the feature selection process. Some experiments provided outstanding results that outperform base-line solutions, even deep learning solutions. Incurred experiments reveal appropriateness of the proposed approach to handle social spam detection problem, providing a trade off between prediction performance and computation capabilities.
Furthermore, the proposed approach is still applicable to a wide range of data mining problems. Below are the key contributions of this research:
\begin{enumerate}
    \item Proposing a social spam content-based detection approach that considers wide variety of contemporary ways of spammers.
    \item Developing a novel genetic algorithm to initialize a powerful classifier and  feature selection.
    \item Validating the proposed approach against publicly released twitter real-world dataset.
\end{enumerate}
\par
The rest of this paper is organized as follows, Section~\ref{Sec:RelatedWorks} investigates literature and related works. The proposed social spam detector and the corresponding genetic algorithm feature extraction approach are elaborated in Section~\ref{Sec:Method}. Results are presented and discussed in Sections~\ref{Sec:Exp} and~\ref{Sec:Disc}, respectively.Finally, Conclusion and future directions are drawn in Section~\ref{Sec:Con}.

\section{Related Works}~\label{Sec:RelatedWorks}

Recently, there has been a significant interest in detecting twitter spam. Compared to the traditional mail spam and web spam, twitter went beyond phishing, fraudulent, and scam. It creates new avenues for profanity, insulting, spreading hate speech, and bullying \cite{kaddoura2022systematic,Article:Chowdhury20,Article:Rubaiee16,article:Liu21,Misc:utkmi_TwitterCompitition19}. Researchers have investigated wide range of approaches to accommodate such divergence. Two streams of countermeasures have been proposed. First approach considers feature extraction. Second approach considers graph-based solution. Feature based solutions investigate content-based features, account-based features, or both of them. Graph based solutions investigate communication graphs of spam spreading  focusing on identification of spammers.
\par


In recent times, feature selection is one of the important key of research in machine learning, image retrieval, text mining, intrusion detection, etc.
According to literature, different algorithms have been developed and employed for feature selection.
For example, a greedy search based sequential forward selection (SFS) \cite{whitney1971direct} and sequential backward selection (SBS) \cite{marill1963effectiveness} have been applied for feature selection. However, these approaches suffer from a range of problems, such as
stacking in local optima and high computational cost.
In order to address these problems, new algorithms for feature selection have been proposed \cite{fogel1998evolutionary,taherdangkoo2013efficient,zhang2007clustering}, such as Particle Swarm Optimization (PSO) \cite{kennedy1995particle}, Ant Colony Optimization(ACO) \cite{dorigo2006ant}, and Genetic Algorithm (GA) \cite{leardi2000application}.
Furthermore, a novel filter feature selection method named the Proportional Rough Feature Selector (PRFS) has been proposed in \cite{wei2020novel}. The method addresses a high dimensional matrix in a short text classification problem. The method makes a regional distinction using a set of terms in order to differentiate documents that exactly belong to a class and documents that possibly belong to a class.
In the work of \cite{liu2008feature}, the authors have presented a comparative study of eight filter methods by employing mutual information using 33 datasets. Furthermore, in the work of \cite{10.5555/944919.944974}, 12 feature selection methods are compared on text classification problem. 

Genetic algorithm has been known to be a very efficient and useful approach for feature selection, as described in \cite{fraser1970computer,oh2004hybrid,baudry2005automatic,akbari2011multilevel,hadizadeh2013quantitative}. This is because of its ability of changing the functional configuration in order to improve the performance results.
In \cite{babatunde2014genetic}, the authors have applied a Genetic Algorithm in order to reduce the number of features extracted from a Flavia image dataset.
The authors of \cite{huang2007hybrid, iqbal2019hybrid} have proposed a hybrid Genetic Algorithm for feature selection based on machine learning techniques. They have investigated the performance of their algorithm using different datasets, such as \textit{Wine} dataset and synthetic data sets.
In \cite{benitez2018improved}, an approach for enhancing a classification performance of natural crisis-related Twitter messages has been proposed. In this approach, a Genetic Algorithm has been utilized for feature selection. 
Another study has been proposed in \cite{aalaei2016feature}. The study employs a Genetic Algorithm for feature selection in order to increase a classification accuracy for breast cancer diagnosis.


Different feature selection approaches have been applied on many real-world applications, such as text categorization \cite{kumar2013mood}, image retrieval \cite{dy2003unsupervised}, intrusion detection \cite{lee2000adaptive}.
Several feature selection approaches have also been applied on tweets classification.
For example, the work of \cite{utama2019sentiment} has presented a method for sentiment analysis of airline tweets. It employs a mutual information method for the process of feature selection. 
Furthermore, the work of \cite{benitez2018improved} has implemented an improved Genetic Algorithm for disaster preparedness and response in the Philippines. The algorithm aims to select the most important features from a large number of features for the classification process of disaster-related tweets.
In \cite{prusa2015impact}, the authors have considered Chi-Squared, Mutual Information, Kolmogorov-Smirnov statistic, area under the Precision-Recall curve, and area under the Receiver Operating Characteristic curve for feature selection on a large high-dimensional dataset of collected tweets. Each tweet is labeled to a positive sentiment or negative sentiment.
The results demonstrated that employing these feature selection techniques on a sentiment classification process can have a great impact on the performance of a classifier.
The \cite{jain2019sentiment} has applied classification techniques on tweets belonging to Renewable Energy. The Correlation based Feature Selection (CFS) Subset Evaluation and Information Gain feature selection have been used to reduce the number of used features.


The literature shows that the number of selected features used for tweets classification greatly affects the performance of the employed classifier. 
However, only a few works have discussed how and what an appropriate number of features should be selected to achieve the best classification performance \cite{kaddoura2022systematic,wang2020optimal}.
The approach of \cite{prusa2015impact} has shown that using between 75 and 200 features enhances the tweets classification results over using the full feature set.
In \cite{saif2012alleviating}, the authors have investigated the using between 42 and 34,855 features to represent 1000 instances from the Stanford Twitter Corpus. They have found that using more than 500 features will not significantly improve the performance of a classifier. 
The work of \cite{suchetha2019comparing} has studied the effect of the application of two-stage feature selection on the twitter sentiment analysis performance. A filter feature selection based on information gain has been used and 3 feature sets of 500, 1000, and 1500 features have been produced.

\section{The Proposed Approach}
\label{Sec:Method}
Computer-based Genetic Algorithm (GA) \cite{holland1992genetic} is a search heuristic that was inspired from the natural evolution theory of Darwin. Since decades, GA has been actively used by researchers to address many challenges in different domains such as malware detection \cite{9422719}, energy optimization \cite{JEYARANJANI2022100710}, cancer classification \cite{8733987} and so on. 

Recently, GA has been used as a search strategy for dimensionality reduction of a relatively large feature space \cite{AMINI2021114072}. Such an approach evades the limitations of the exhaustive search strategies. GA can be used to optimize the parameters of machine learning algorithms and reduce the dimensionality of the problem space.
One approach in text modeling is to convert the input text into a set of features; such as TF/iDF modeling. Usually, the number of features is extremely large and so an overfitting probability is high. On the other hand, real-world classification datasets are usually an imbalanced distribution of class labels. Imbalanced datasets impose an additional challenge in avoiding classification bias and overfitting.

\subsection{eXtreme Gradient Boosting}

\begin{figure}[h]!
	\centering
		\includegraphics[scale=0.25]{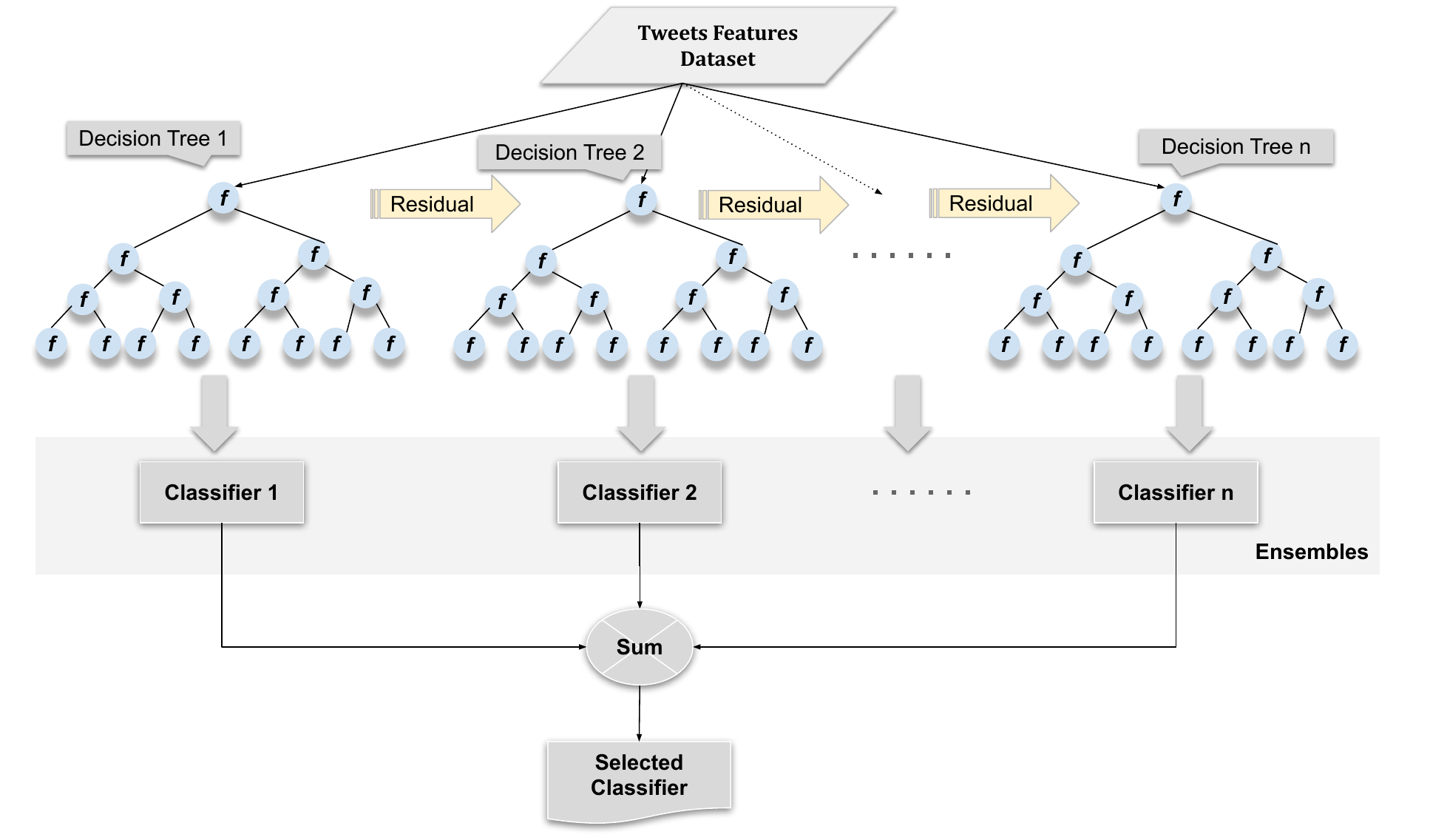}
	\caption{ eXtreme Gradient Boosting (XGBoost)}
	\label{FIG:XGBoost}
\end{figure}

Chen and Guestrin in \cite{Chen:2016:XST:2939672.2939785} introduced a powerful tree boosting algorithm, which is named eXtreme Gradient Boosting (XGBoost). The algorithm is claimed to be scalable; sparsity-aware; takes into consideration data compression and sharding; and cache-aware access. Figure \ref{FIG:XGBoost} illustrates XGboost algorithm architecture. Each tree is trained on the residual error of the previous tree which improves the performance of the constructed tree. The sum of each tree's predictions constructs the final prediction.

The characteristics of XGBoost enable it to outperform other machine learning algorithms and require less system resources. Theoretical and empirical proofs support these claims in \cite{OCCHIPINTI2022117193,Chen:2016:XST:2939672.2939785,benchm-ml2019,morde_setty_2019,nielsen2016tree}. XGboost in \cite{OCCHIPINTI2022117193} produced the best performing models over 11 machine learning algorithms in text-based spam classification. However, tree-based algorithms in general tend to perform well in relatively small number of features compared to artificial neural networks. Therefore, this research aims at leveraging the benefits of XGBoost algorithm by reducing the number of text features in the prediction model building process.  

The major challenge in building XGBoost-based models is proper parameter tuning \cite{OCCHIPINTI2022117193,Chen:2016:XST:2939672.2939785,benchm-ml2019,morde_setty_2019}. This research aims at proposing a novel GA variation that optimizes the parameters of a classifier (i.e., eXtreme Gradient Boosting), and to reduce the features space simultaneously.

\subsection{Dataset Description}
\begin{figure}[h!]
	\centering
		\includegraphics[scale=0.6]{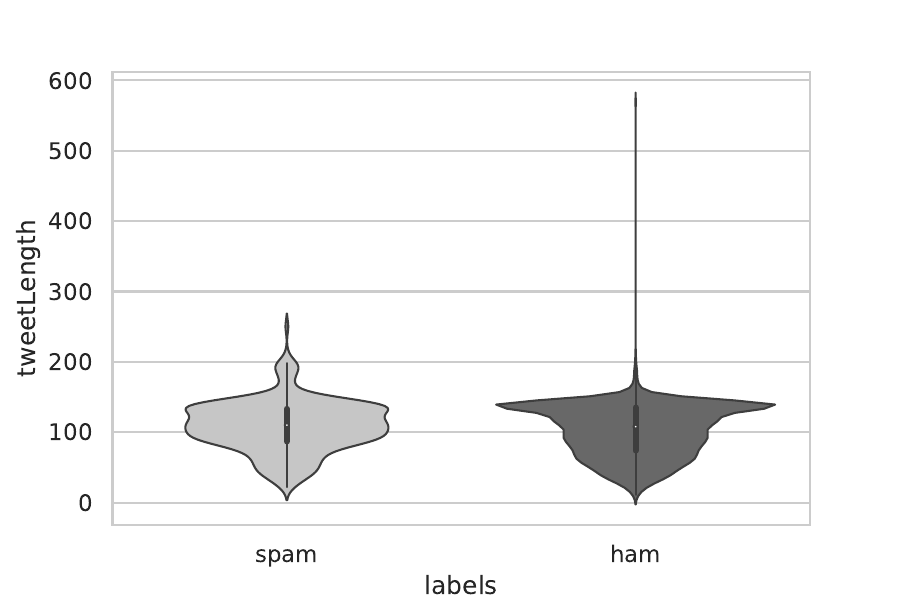}
	\caption{ The character length distribution of the text (emoticons are stored as a series of Unicode characters).}
	\label{FIG:tweetLenBoxP}
\end{figure}

The main tweets dataset used in this research was introduced in \cite{jain2019optimizing}. It has 5096 tweets. About 17\% of tweets are labeled as ``Spam'' and the rest as ``Ham''. Tweets are labeled in a manual fashion by considering and examining each one separately \cite{jain2019optimizing}. If a tweet content is considered as unacceptable by the community or harmful,then the tweet is labeled as spam. Otherwise, it is labeled as ham (i.e.,, normal tweet). Figure \ref{FIG:tweetLenBoxP} summarizes the number of the text characters in all instances. The length of the stored characters of each tweet may exceed the number of original tweet length because some special characters and emoticons are stored as a set of representative Unicode characters.

It is apparent that the average character length, which represents each tweet, is about 100 characters in both classes, and there is no significant variance difference in the distribution of both classes as well. Therefore, length analysis adds to the challenges in building a robust classification model.

\subsection{The Modified GA}
The modified GA aims at directing the stochastic selection aspect towards a fine subset of features. At the same time, to find the best possible classification algorithm parameters. Therefore, it is to find the best combination of features and parameters simultaneously. Usually, GA is used to either initialize the classification algorithm parameters or in feature selection. The proposed modifications would leverage the capabilities of GA in defining the optimal combination of features subset and parameters. Moreover, particular modifications of some methods limit the absolute randomness of GA phases. For example, ensuring no duplicate genes in each chromosome. The Modified GA and its phases are presented in the following subsections:

\subsubsection{GA main code}

Many recent research studies in different domains  \cite{PAN2022109520,THITHUYLINH20223301,kavzoglu2022advanced,deng2022hybrid} illustrated the power of GA in optimizing the parameters of XGBoost to achieve better prediction performance. Algorithm \ref{algGA} represents the initial configurations of the modified GA that is used to optimize the parameters of XGboost and select the most appropriate features subset. Initially, a number of GA parameters will set the maximum percentage of features to be selected, the parents' crossover ratio in each population, the maximum number of generations, and the number of classifier parameters to be optimized. The result will be a chromosome having an optimized set of XGboost parameters and the selected features subset. The chromosome structure is illustrated in Figure \ref{FIG:Chromosome_CrossMut}. The input dataset is split into 70\% training and 30\% testing partitions for the GA-based XGBoost model building and validation. Table \ref{tbl:gaparameters} describes the GA parameters. 

\begin{algorithm}
\caption{Genetic Algorithm}\label{algGA}
\begin{algorithmic}[1]

\State Set percentage of features to be selected 
\State Set crossover ratio 
\State Set number of parents in initial population
\State Set number of generations
\State Number of parents to select = crossover ration * number of parents
\State Define classifier parameters to be optimized 

\State generateTrainTestData(Features Dataset) \Comment{30:70\% data split}

\State Generate initial population(\texttt{Number of parents})
\State fitness function(initial generation)

\For{\texttt{Number of generations}}
\State train(current population)
\State Select parents (Number of parents to select) \Comment{return those having highest fitness value}
\State children = crossover(selected parents)
\State mutate (children)
\EndFor

\State Best Chromosome = highest fitness chromosome


\Procedure{train}{current population}
\For{\texttt{c = 1 to Number of chromosomes}}
\State chromosome = population[c] 
\State $parameters\gets chromosome[1-7]$
\State $features\gets chromosome[8-end]$
\State predictions = classify(parameters, traindata[features])
\State fitness function(actual labels, predictions)
\EndFor
\State \textbf{return} $fitness$
\EndProcedure


\end{algorithmic}
\end{algorithm}

\subsubsection{Initial population}

Creating the initial population of the GA is challenging as it is not an easy task to select a representative subset of the whole population. Neither in selecting the initial set of classifier parameters nor the subset of the feature set. Redundancy of gene values is also one of the issues to consider at this phase. To limit the absolute randomness of the GA, XGBoost boosting parameters are generated using a uniform random number generator within a recommended value range (Table \ref{tbl:xgboostparams} lists the value ranges of XGBoost parameters).

The features subset, which is part of the chromosome, is created by a custom procedure that randomly selects a subset of the whole feature set; i.e., subset of the whole TF-iDF vector. The procedure ensures creating a chromosome with no duplicate features and selecting from the full features vector. Actually, the list of selected features is the set of features indices in the TF-iDF vector; 

Initializing the initial population will result in forming the parents chromosomes according to the preset parents size in Algorithm \ref{algGA}.

\begin{algorithm}
\caption{Initializing initial population}\label{algInitPop}
\begin{algorithmic}[1]
\Procedure{Generate initial population}{$number of parents$}
\For{p = 1 to number of parents}
\State learningRate[p] = rand.uniform(0.01, 1)
\State nEstimators[p] = randrange(10, 1500, step = 25)
\State maxDepth[p] = randrange(1, 10, step= 1)
\State minChildWeight[p] = rand.uniform(0.01, 10.0)
\State gammaValue[p] = rand.uniform(0.01, 10.0)
\State subSample[p] = rand.uniform(0.01, 1.0)
\State colSampleByTree[p] = rand.uniform(0.01, 1.0)
\State features[p] = select text features
\Comment{select text features ensures no duplicate genes are present in each chromosome}
\EndFor
\State concatenate parameters and features into chromosomes
\State population = all generated chromosomes
\State \textbf{return} $population$
\EndProcedure
\end{algorithmic}
\end{algorithm}

\subsubsection{Fitness function}

The bias imposed by imbalanced class distributions generally favors the majority class; which in most cases does not represent the class of interest. Therefore, positive class based metrics will dramatically mislead the selection of the best model relying on the objective function. The Geometric Mean (GMean) on the contrary considers both the positive and negative class as an objective function \cite{4479477,kim2015geometric,akosa2017predictive}. The GA and the validation of the selected models in this research utilize the GMean as an objective function. In addition, it is used as the main metric in comparing the performance of different classification models. The Algorithm \ref{algFitFun} and Equation \ref{eq:gmean} illustrate the GMean calculation. It is the square root of the True Positive Rate (TPR or Recall) multiplied by the True Negative Rate (TNR or Specificity). The TPR is a positive class based metric and TNR is a negative class metric; deriving a TPR and TNR based metric equals a metric that represents the accuracy of both classes (i.e., Spam and Ham).

\begin{algorithm}
\caption{Simplified fitness function}\label{algFitFun}
\begin{algorithmic}[1]
\Procedure{fitness function}{$y\_true, y\_pred$} \Comment{y\_true is actual class labels, y\_pred predicted labels}
\State TP = Count TP($y\_true, y\_pred$)
\State TN = Count TN($y\_true, y\_pred$)
\State FP = Count FP($y\_true, y\_pred$)
\State FN = Count FN($y\_true, y\_pred$)
\State \texttt{$TPR = \frac{TP}{TP+FN}$}
\State \texttt{$TNR = \frac{TN}{FP+TN}$}
\State \texttt{$fitness = \sqrt{TPR \times TNR}$}
\State \textbf{return} $fitness$
\EndProcedure
\end{algorithmic}
\end{algorithm}


\subsubsection{Selection and Crossover}

Crossover is an essential phase in GA to generate a new number of children from the parents. A child's genes will be a combination of two parent chromosomes, so the children are expected to have better genes than the parents do. To achieve this, a uniform crossover is performed using almost half of each parent genes. The crossover phase ensures generating children where each has no duplicate genes. Each chromosome will undergo two crossovers; one for XGBoost parameters and the other for the selected features set. Algorithm \ref{algSelection} will select the best parents to crossover according to the GMean objective function and generate a number of new children for the next generation.

\begin{algorithm}
\caption{Crossover}\label{algSelection}
\begin{algorithmic}[1]
\Procedure{crossover}{Selected parents} \Comment{Uniform crossover}
\For{number of children to generate}
\State uniform crossover of selected parents parameters
\State uniform crossover of selected parents features
\State new child = parents parameters and features after crossover
\EndFor
\State \textbf{return} $newChildren$

\Comment{ensures no duplicate genes are present in each child chromosome}

\Comment{ensures uniform crossover at almost half the size of each parent}

\Comment{first 7 genes are classifier parameters and the rest are the text features}

\EndProcedure
\end{algorithmic}
\end{algorithm}

\subsubsection{Mutation}
Because of the stochastic nature of GA, some genes may be overseen in the initial population or in the generated children. To increase the chance of fair inclusion of missed geneses the mutation tries to include new genes in the children. One parameter gene and one feature gene are selected randomly in each child chromosome and replaced with a new value. Specifically for the selected features, the mutated gene value will be selected from the full features vector set such that it is not one of the parents' genes. Mutation is illustrated by Algorithm \ref{algMutation}.

\begin{algorithm}
\caption{Mutation}\label{algMutation}
\begin{algorithmic}[1]
\Procedure{mutate}{children}
\For{number of children}
\State randomly select parameter gene index
\State add random value to existing gene value
\State Ensure not exceeding parameter value range 

\State randomly select feature gene index
\State mutate by replacing it with a value from the full features set
\State ensure adding new feature to the current chromosome
\EndFor
\State \textbf{return} $MutatedChildren$
\EndProcedure
\end{algorithmic}
\end{algorithm}

The modified GA will run for a preset number of generations aiming at the maximization of the GMean value of the generated models. The spam and ham text features are the TF-iDF vectors generated by the pre-processing step presented in Section \ref{DSpreprocessng}. The best chromosome that will be selected in the last GA generation will contain the best XGBoost parameters and the accompanying selected spam features subset. This chromosome will be used consequently to initialize an XGBoost algorithm to generate a spam prediction model in the Model Building phase illustrated in Figure \ref{FIG:Methodology_Abstract} and Figure \ref{FIG:Methodology_OptFSFA}. 

\subsection{Proposed Methodology}
\label{propMethodology}
\begin{figure}[H]
	\centering
		\includegraphics[scale=.4]{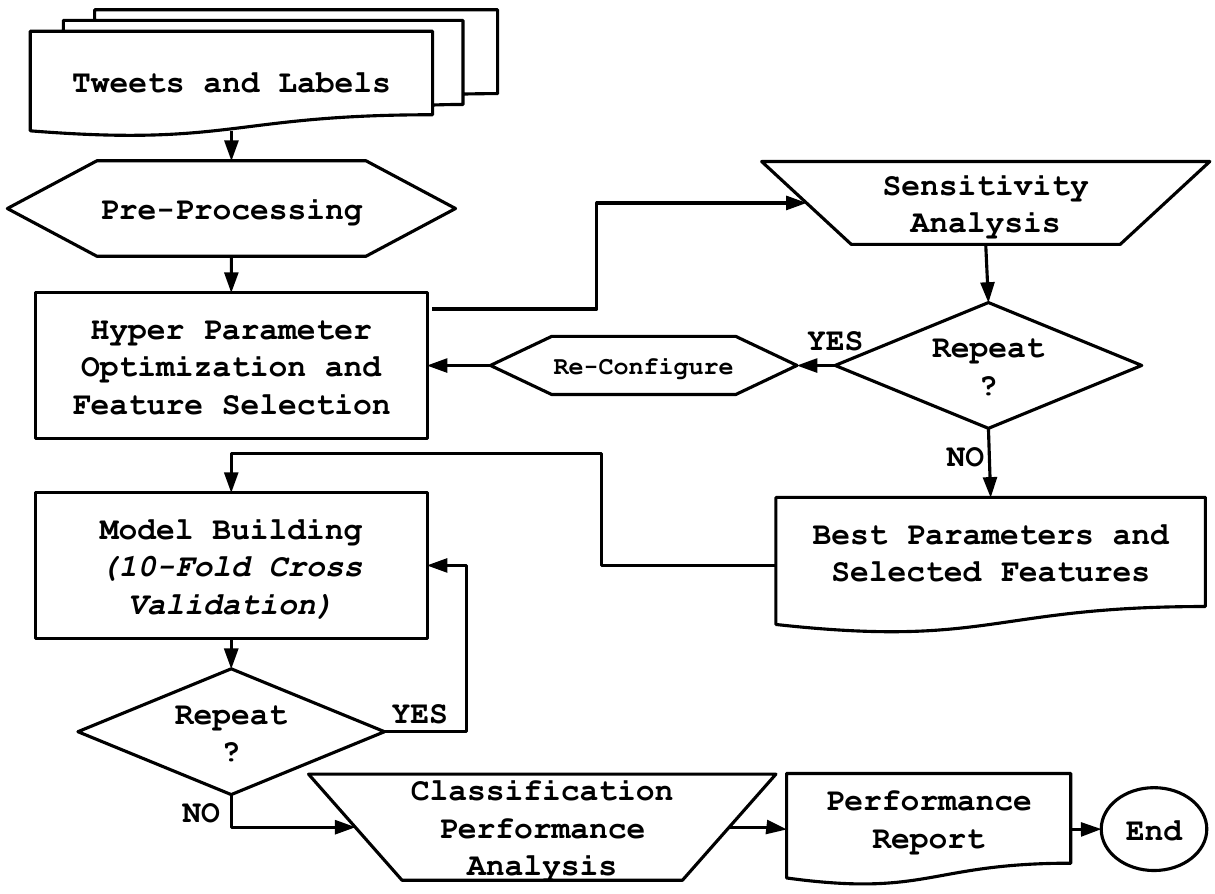}
	\caption{ The proposed research methodology.}
	\label{FIG:Methodology_Abstract}
\end{figure}

\begin{figure*}[h]
	\centering
		\includegraphics[scale=.5]{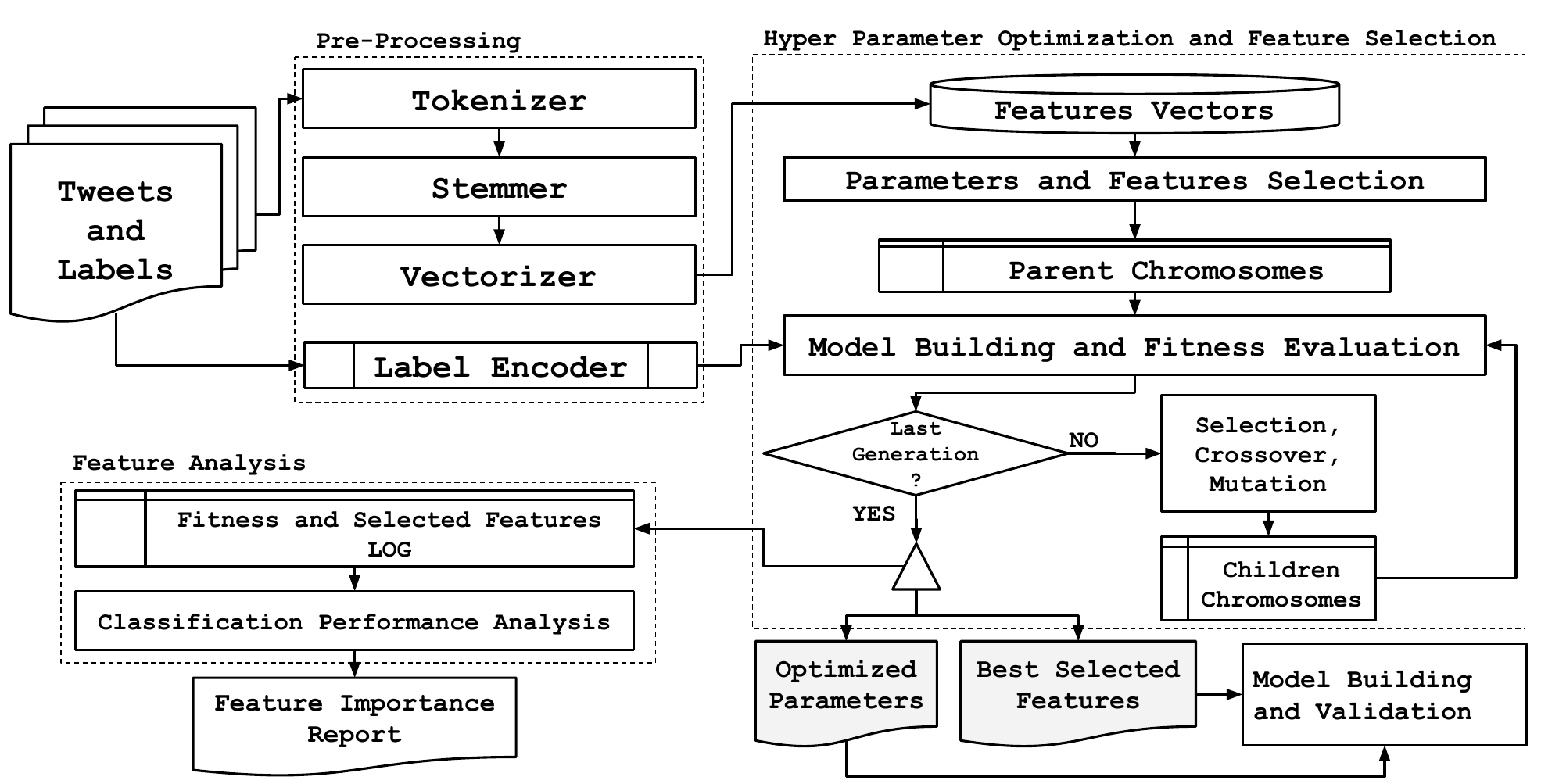}
	\caption{ The detailed research methodology.}
	\label{FIG:Methodology_OptFSFA}
\end{figure*}

The proposed methodology is divided into five main phases: (1) Dataset pre-processing, (2) Hyper parameter optimization and feature selection, (3) Sensitivity analysis, (4) Model building and validation, and (5) Classification performance analysis. Figure \ref{FIG:Methodology_Abstract} is an abstract view of the proposed research methodology. Figure \ref{FIG:Methodology_OptFSFA} is a more detailed view of the methodology, and the parameter values of each step are listed in Section \ref{Sec:Exp}.

\subsubsection{Dataset Pre-processing}\label{DSpreprocessng}
Each data instance contains raw tweet text and a label (i.e., ``Ham'' or ``Spam''). Each tweet text and its label is pre-processed to be cleaned and converted into features through a number of steps: (1) Tokenize the tweet and remove extra space and special characters, (2) Stem each tokenized word using ``Porter Stemmer'' \cite{porter1980algorithm}. This will reduce the tokenized word to its root, stem, or base.  (3) Each stem is given a weight using a vectorizer; depending on the term frequency–inverse document frequency (TF-iDF) \cite{Salton:86,scikit-learn}. Therefore, each tweet is converted into a representative TF-iDF vector (i.e., a set of features), and (4) The class labels are encoded into 0's, i.e.,, ``Ham'' class, and 1's, i.e.,, ``Spam'' class, to satisfy the requirements of the classification algorithm.  

\subsubsection{Hyper parameter optimization and feature selection}
\begin{figure}[H]
	\centering
		\includegraphics[scale=.28]{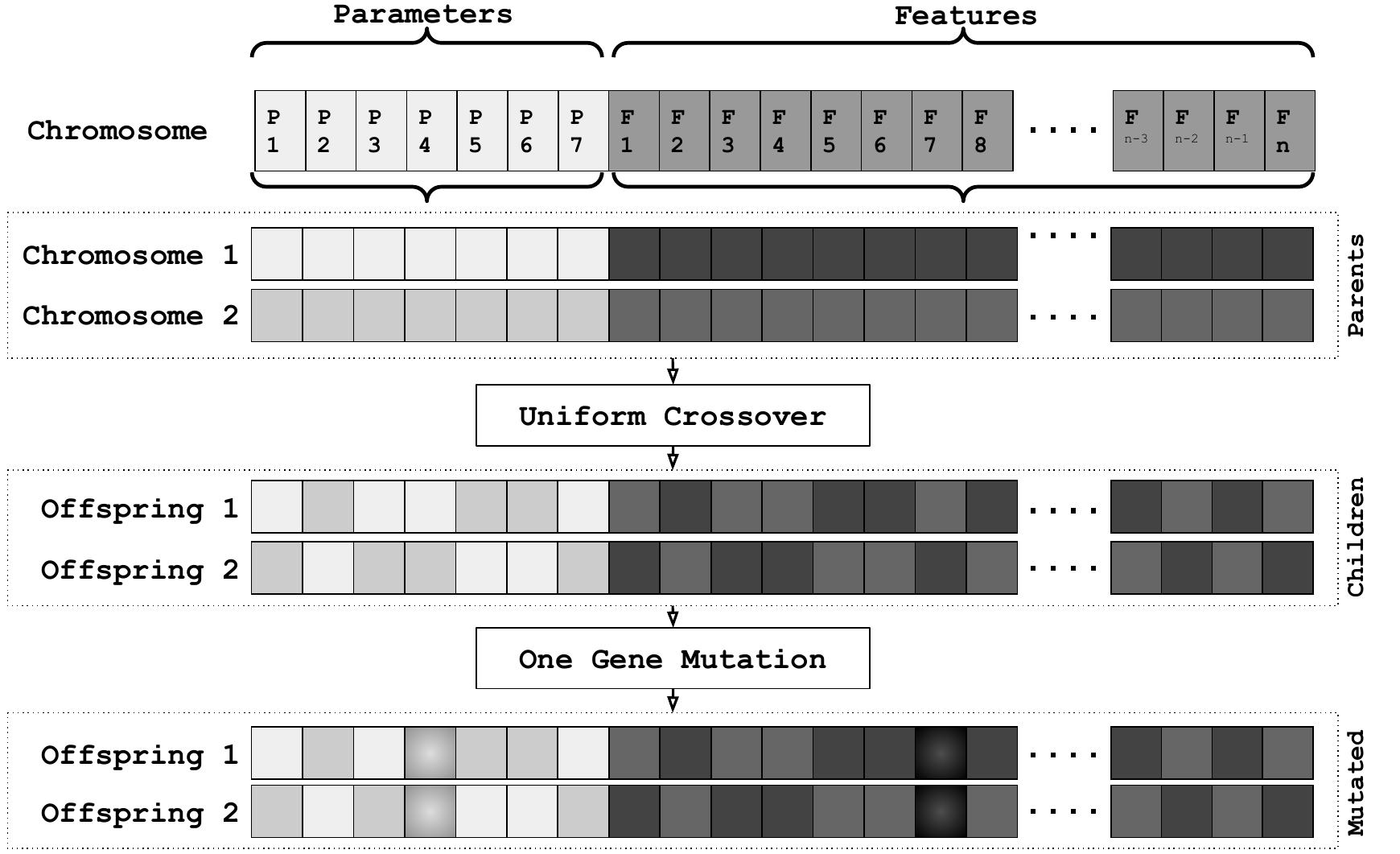}
	\caption{ The chromosomes structure, crossover, and one gene mutation.}
	\label{FIG:Chromosome_CrossMut}
\end{figure}

In this step, a modified GA tunes the parameters of the classification algorithm such that it improves the prediction rates. It is divided into two main parts: (a) GA feature selection and (b) GA hyper parameter optimization. Each chromosome in this step is designed to hold two types of genes such that genes at the beginning are the parameters to be optimized and the rest of genes are the selected features. Figure \ref{FIG:Chromosome_CrossMut} shows the detailed structure of the chromosomes, the chromosomes after GA crossover, and an illustration of the genes after the mutation process.

The initial population consists of parent chromosomes holding randomly selected parameters within a recommended and pre-defined range based on literature  \cite{Chen:2016:XST:2939672.2939785,chen2020improved,jiang2019pedestrian}), and randomly selected unique features within each chromosome.  It is the responsibility of the initialization algorithm to ensure choosing features without having any duplicates in each chromosome.

\subsubsection{Sensitivity analysis}\label{sub:SA}
The main aim of the sensitivity analysis step is to find the best possible combination of XGBoost parameters and subset of feature space \cite{saltelli2002sensitivity}. GA performs several hyper parameter optimizations and feature selections in order to examine the behavior of the classification algorithm. Consequently, the results of different optimizations and feature selections under different configurations lead to understanding the behavior of the algorithm.

Several GA configurations are examined in this step by mainly specifying: (1) the initial population size, crossover percentage, and number of generations; and (2) the desired percentage of features to retain (i.e., the number of feature genes in the chromosomes). The effect of the configurations on the objective function is examined to determine the candidate classifier parameters and the subset of feature space.

\subsubsection{Model building and validation}
The optimized classifier parameters and selected features subset, which maximized the objective function, are used to build a robust classification model. 10-Fold stratified cross-validation (10 CV) is used to avoid the bias in model building process. This model building process is repeated 50 times ($50\times10$ CV) to assess the classifier stability.

\section{Experiments Setup}
\label{Sec:Exp}
Google\texttrademark\ Colaboratory (a.k.a. Colab, \url{https://colab.research.google.com/}) environment is one of Google\texttrademark\ Research products. Colab offers a browser-based machine learning projects' development environment that supports Python\texttrademark\ \url{https://www.python.org/} code run over different modern processing architectures. The experiments of this research are implemented as Python\texttrademark\ projects and conducted over CPU-based Colab environment.       

The major configurations of the proposed approach steps, which are described in Section \ref{propMethodology}, are listed as follows:

\subsection{Dataset Pre-processing}
The used TF-iDF vectorizer parameters in Tweets text pre-processing are listed in Table \ref{tbl:tfidfvectparams}. The maximum number of possible features is extracted according to the pre-processing step.

\begin{table}[H]
\caption{TF-iDF vectorizer parameters.}
\label{tbl:tfidfvectparams}
\centering
\setlength{\tabcolsep}{17pt}
\begin{tabular}{lc}
\toprule
Vectorizer Property     & Value \\
\midrule
Regular Expression Processing & r``{[}\textasciicircum{}A-Za-z0-9\textbackslash{}-{]}\textbackslash{}@*'' \\
Stemmer                 & Porter               \\
Stop Words              & Not filtered         \\
Max features to extract & Extract all features \\
Encoding                & ISO-8859-1  \\
\bottomrule
\end{tabular}
\end{table}

\subsection{GA Configuration}
Tables \ref{tbl:gafilenamecodes} and \ref{tbl:gaparameters} show the GA parameters setup. The letters $F$, $P$, $C$, and $G$ are used to summarize the description of each GA configuration (i.e., Metadata elements). $F$ is the percent of features subset to be selected from the complete features set, $P$ is the number of randomly selected parents in the first GA generation, $C$ is the number of parents to crossover, and $G$ is the number of GA generations. This standard file naming convention makes it easier to sort and interpret some effects of parameter tuning.

\begin{table}[H] 
\caption{GA file-name/experiment code description.}
\label{tbl:gafilenamecodes}
\centering
\begin{tabular}{p{0.3cm}p{3.4cm}p{2.9cm}}  
\toprule
Code & Description  & Values                              \\
\midrule
F    & Percent of \textbf{F}eatures to select  & 1,5,10,20,30,40 \%     \\
P    & No. \textbf{P}arents (First generation) & Rough tuning values\\
C    & No. parents \textbf{C}rossover &   Based on different crossover ratios \textit{(the range for the first ten experiments: 10 to 100\%)} \\
G    & No. \textbf{G}enerations &  100 for the first ten experiments, then 50 ,and for very large P values it was set to 5            \\
\bottomrule
\end{tabular}
\end{table}

\begin{table}[H] 
\caption{GA parameters description.}
\label{tbl:gaparameters}
\begin{tabular}{p{0.4\linewidth}  p{0.5\linewidth}}
\toprule
GA Parameter          & Description         \\
\midrule
Perc2Sel              & Percent of parents to select from the features set   \\
CrossoverRatio        & Percent of parents' crossover rate \\
NumFeat2Select        & Derived value, depends on Perc2Sel to define the number of features to be selected      \\
numberOfParentsMating & Derived value, depends on CrossoverRatio to define the number of parents in crossover     \\
numberOfXGBParameters & Number of classification algorithm parameters to be optimized (i.e., 7)   \\
numberOfParameters    & Chromosome size (i.e.,Total number of parameters to be optimized and max number of  features to be selected which equals $numberOfXGBParameters + NumFeat2Select$) \\
numberOfGenerations   & The number of generations that will be created\\
\bottomrule
\end{tabular}
\end{table}


The best results of GA feature selection and hyper parameter optimization are selected based on fitness function (i.e., GMean); which consist of the selected features and optimized XGBoost parameters, which are listed in Table \ref{tbl:xgboostparams}).


\begin{table*}[h] 
\caption{XGBoost boosting parameters to be optimized by the modified GA}
\label{tbl:xgboostparams}
\centering
\begin{tabular}{lcl}   
\hline
XGB Parameter  & Value Range     & Description \\
\hline
learning\_rate (eta) &  0.01 - 1  & Algorithm learning rate; lower the better but requires more iterations to find optimal solution.     \\
n\_estimators      & 10 - 1500 &  Maximum number of estimators         \\
max\_depth   &   1 - 10    & Maximum tree depth, to control overfitting. (e.g.,high depth will biase the algorthim towards a specific sample)         \\
min\_child\_weight &  0.01 - 10.0  &  Minimum sum of observations in a child         \\
gamma     &    0.01 - 10      & Minimum reduction of loss when splitting        \\
subsample     &  0.01 - 1.0    & Random subset of observations for each tree           \\
colsample\_bytree & 0.01 - 1.0 & Subset of columns to be samples in the trees         \\
seed               & Fixed at 723 & Used in parameter tuning and to have reproducible  results \\
\hline
\end{tabular}
\end{table*}

        
\subsection{Classification performance analysis}

Analyzing the performance of the classifier to demonstrate its learning capability in model development is an essential part of the modeling phase assessment. Therefore, several metrics illustrate the performance of the developed model in detecting a potential spam tweet (i.e., classifying the tweets into Spam and Ham).  

A visual summary of the classification results is represented by a confusion matrix \cite{SammutWebb2017}. Such that a two dimensional table aggregates the counts of the labeled tweets by the developed classification model into correct (True) and incorrect (False) labels. The aggregated counts are denoted specifically as True Positive ($TP$), False Positive ($FP$), True Negative ($TN$), and False Negative ($FN$). In this work, the positive class (i.e., class of interest) is the Spam tweet, and the negative class is the Ham tweet. Consequently, the four aggregated counts in the confusion matrix are interpreted as follows: $TP$ count represents the actual Spam tweets that are classified correctly as Spam, $TN$ count is the number of actual Ham tweets that are classified correctly as Ham, $FP$ count is the number of the actual Ham tweets that are classified incorrectly as Spam, and the number of actual Spam tweets that are classified incorrectly as Ham represents the $FN$ count. $TN$ and $TP$ represent the goodness of the classification model in correctly predicting the class label, while $FP$ and $FN$ show the level of the possible confusion a prediction model may have. Table \ref{tbl:confusionMatrix} represents the confusion matrix that is used to derive a number of Spam classifier performance evaluation metrics.

\begin{table}[H]
\caption{Confusion~Matrix.}
\label{tbl:confusionMatrix}
\centering
\setlength{\tabcolsep}{18pt}
\begin{tabular}{lccc}
\toprule
\multicolumn{2}{c}{} & \multicolumn{2}{c}{\begin{tabular}[c]{@{}c@{}}Predicted Tweet Class\end{tabular}} \\ \midrule
\multicolumn{2}{c}{} & \multicolumn{1}{c|}{Spam} & \multicolumn{1}{c}{Ham} \\ \cmidrule{2-4} 
\multicolumn{1}{l}{\multirow{2}{*}{\begin{tabular}[c]{@{}l@{}}Actual Tweet\\ Class\end{tabular}}} & \multicolumn{1}{c}{Spam} & \multicolumn{1}{c}{$TP$} & \multicolumn{1}{c}{$FN$} \\ \cmidrule{2-4} 
\multicolumn{1}{l}{} & \multicolumn{1}{c}{Ham} & \multicolumn{1}{c}{$FP$} & \multicolumn{1}{c}{$TN$} \\ \bottomrule
\end{tabular}
\end{table}

Part of the derived evaluation metrics are:
\begin{enumerate}
\item True Positive Rate ($TPR$): the ratio of the correctly classified Spam tweets (i.e., tweets predicted as Spam and they are actually a Spam) \cite{SammutWebb2017}. It is alternatively named as recall or sensitivity.
\begin{equation}
TPR=\frac{TP}{TP+FN}
\end{equation}

\item True Negative Rate ($TNR$): the ratio of the correctly classified Ham tweets (i.e., tweets predicted as Ham and they are actually Ham) \cite{SammutWebb2017}. It is alternatively named as specificity.
\begin{equation}
TNR=\frac{TN}{FP+TN}
\end{equation}



\item Positive Predictive Value ($PPV$). It is alternatively named as precision \cite{SammutWebb2017}:   
\begin{equation}
PPV=\frac{TP}{TP+FP}
\end{equation}

\item False Positive Rate ($FPR$): the probability of false alarm, Fall-out.
\begin{equation}
FPR=\frac{FP}{FP+TN}
\end{equation}

\item Negative Predictive Value ($NPV$) \cite{SammutWebb2017}:
\begin{equation}
NPV=\frac{TN}{TN+FN}
\end{equation}

\item F-Score ($F1$) \cite{SammutWebb2017}:
\begin{equation}
F1= 2 \times \frac{TPR \times PPV}{TPR + PPV}
\end{equation}

\item Total Accuracy ($Accuracy$): It is traditionally derived from the confusion matrix and it represents the correctly classified instances count divided by the total number of instances. Alternatively, accuracy is also referred to as success rate (i.e.,, the ratio of correctly classified instances). Equation \ref{Eq:acc} illustrates accuracy metric.
\begin{equation}
Accuracy =\frac{TP+TN}{TP+TN+FP+FN}
\label{Eq:acc}
\end{equation}

\end{enumerate}

There are many concerns in using the total accuracy as a performance metric, more particularly in imbalanced datasets \cite{chawla2009data,ghatasheh2020cost,ghatasheh2020business}. Usually the negative class is dominant and more frequent in real life. Consequently, the model building phase would have a higher tendency towards modeling better the patterns of the negative class. Such tendency makes less prediction power of the positive class; which is usually the class of interest. Same issue arises when considering Spam tweets. While TNR tends to rise up, TPR tends to decline. Therefore, further evaluation metrics are advised here.  

The Geometric Mean ($GMean$) and Area Under the Curve ($AUC$) are used commonly in evaluating the classifiers of imbalanced class distribution. GMean and AUC take into consideration the minority class and seek the balance between the classes in illustrating the model accuracy (i.e., class independent metrics). The GMean is calculated according to Equation \ref{eq:gmean}; i.e., the square root of the recall of the positive class multiplied by recall of the negative class. The calculation of the GMean ensures unbiased behavior of the metric either in objective function evaluation or in performance evaluation. A higher GMean value indicates better performance of the classifier \cite{4479477,kim2015geometric,akosa2017predictive}.

\begin{equation}\label{eq:gmean}
GMean=\sqrt{TPR \times TNR}
\end{equation}

Receiver Operating Characteristic Curve ($ROC$) is a metric that takes different threshold values and confront them with the corresponding probabilities (i.e., $TPR$ and $FPR$). The $AUC$ is generated by calculating the area under the $ROC$. Therefore, there is a positive correlation between the value of $AUC$ and the diagnostic ability of the classification model \cite{hanley1982meaning}.

\section{Results and Discussion}
\label{Sec:Disc}
It takes a considerable amount of computation time to have a relatively robust tweets spam prediction model using GA. The proposed methodology is quite complex and strives to find an optimal subset of tweet features and classifier parameters simultaneously. In tweets pre-processing step, the TF-iDF vectorizer generates the maximum number of possible features per tweet along with their TF-iDF value (i.e., the total of 14343 features extracted from all tweets). 

Due to the time complexity of each GA search process, an initial relatively small subset of features (i.e.,, 1\%) is examined in order to study the performance behavior of the classifier and select the most appropriate algorithm configurations (i.e., sensitivity analysis). The performance analysis of the first 10 experiments relies on 1\% of feature space, 10 initial population parents, and 100 generations. Next, the generations are fixed at 50 and the crossover ratio at 60\% of parents. Finally, the effect of different feature subset size is examined (i.e., F=1 , 5, 10, 20, 30, and 40).

A number of selected features and parameters are used to $10 \times 50$ cross-validate prediction models. The performance metrics (GMean in particular) indicate promising capabilities of the modified GA in finding a subset of the feature space and optimizing the parameters of the classifier accordingly. The outcomes of this research are compared to related work in terms of dimensionality reduction. 

It is worth noting that the performance metrics in this research are presented such that further comparison with existing or future research is possible. The following subsections show and discuss the findings in more detail.

\subsection{Sensitivity Analysis}

\begin{figure}[H]
	\centering
		\includegraphics[scale=0.55]{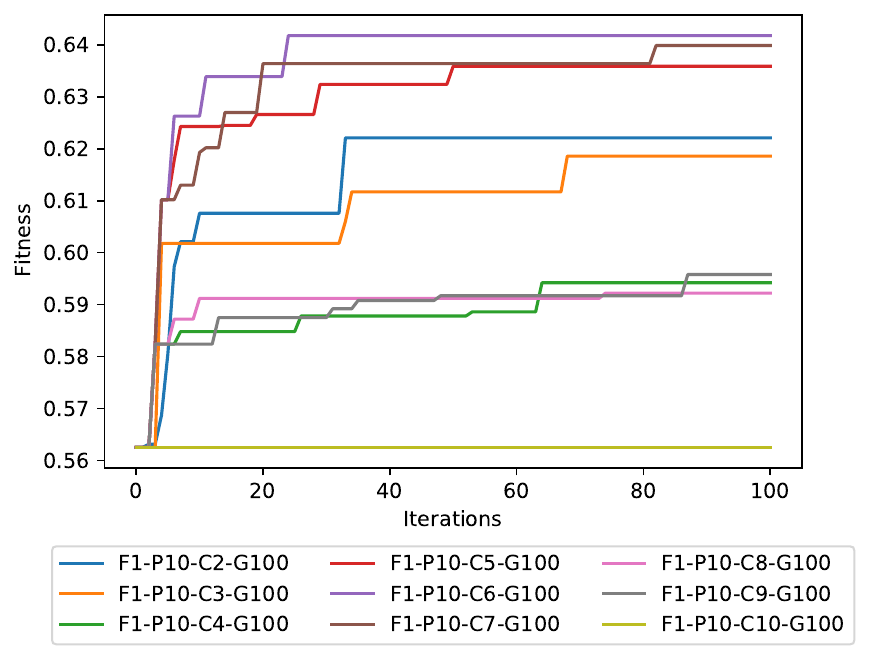}
	\caption{ Sensitivity analysis: Number of iterations and crossover ratio by setting the parameters described in Table \ref{tbl:gafilenamecodes}}
	\label{FIG:FitCurve}
\end{figure}

Sensitivity analysis results show the effect of GA crossover ratio on classification performance and its convergence. This is an important step to predict the behavior of the GA, and define the crossover ratio and number of GA iterations. Figure \ref{FIG:FitCurve} represents the fitness curve of several crossover ratios ranging from 2\% to 10\%; while the remaining three parameters have been fixed at 1\%of the total number of features, 10 parents, and 100 generations.
Almost similar convergence pattern of the algorithm is present but a significant difference in fitness value groups the results into three main levels (i.e., Low, moderate, and high fitness value groups that correspond to different crossover ratios). 
Extreme values of crossover rate lead to very poor fitness values and lower prediction rate. Low values of crossover rate lead to relatively moderate fitness values. Moderate crossover values lead to relatively high fitness values, such that leveraging the prediction power of the spam classification models.
It is apparent that 60\% crossover rate is the most appropriate diversity factor to the next GA population.
Experiments reveal that the number of iterations (i.e., number of GA populations) reaches a relatively local maximum early and slightly increases in consequent generations. Consequently, the number of appropriate generations is lowered to be 50 keeping in mind the effect on larger initial populations and number of features.

Several GA optimization and feature selection experiments (Table \ref{tbl:resultsGA}) aim at maximizing the fitness value. Some of the experiments cross over ratio ($C$) are fixed at 60\%, 50 generations ($G$), and examine several percent of features to select ($F$). Furthermore, Table \ref{tbl:featurefrequencyinexp} lists some of the most frequently selected features in all the experiments and the presence of these top features in the best performing models.

\begin{table*}[h] 
\caption{Fitness value obtained by several GA feature selection and optimization experiments}
\label{tbl:resultsGA}
\centering
\setlength{\tabcolsep}{19pt}
\begin{tabular}{lc|lc|lc}
\toprule
ExpID            & Fitness & ExpID            & Fitness & ExpID              & Fitness \\
\midrule
F1-P10-C1-G50    & 61.43\%     & F1-P250-C150-G50 & 75.19\%     & F1-P50-C5-G50      & 65.02\%     \\
F1-P10-C10-G100  & 56.25\%     & F1-P250-C5-G50   & 66.87\%     & F1-P500-C250-G50   & 81.13\%     \\
F1-P10-C2-G100   & 62.21\%     & F1-P300-C10-G50  & 71.96\%     & F1-P500-C300-G50   & 82.77\%     \\
F1-P10-C3-G100   & 61.86\%     & F1-P300-C100-G50 & 75.99\%     & F1-P500-C5-G50     & 75.96\%     \\
F1-P10-C4-G100   & 59.42\%     & F1-P300-C180-G50 & 77.92\%     & F1-P600-C360-G50   & 81.64\%     \\
F1-P10-C5-G100   & 63.59\%     & F1-P300-C20-G50  & 73.22\%     & F5-P300-C100-G50   & 77.65\%     \\
F1-P10-C6-G100   & 64.18\%     & F1-P300-C30-G50  & 74.68\%     & F10-P2870-C1722-G5 & 78.10\%     \\
F1-P10-C7-G100   & 63.99\%     & F1-P300-C40-G50  & 75.22\%     & F10-P300-C100-G50  & 80.33\%     \\
F1-P10-C8-G100   & 59.22\%     & F1-P300-C5-G50   & 76.64\%     & \textbf{F10-P400-C240-G50}  & \textbf{84.85\%}     \\
F1-P10-C9-G100   & 59.58\%     & F1-P300-C50-G50  & 73.39\%     & F10-P5023-C3014-G5 & 80.22\%     \\
F1-P100-C5-G50   & 65.86\%     & F1-P350-C5-G50   & 73.86\%     & F20-P300-C100-G50  & 83.99\%     \\
F1-P100-C50-G50  & 73.28\%     & F1-P400-C240-G50 & 80.99\%     & F20-P5738-C3443-G5 & 80.69\%     \\
F1-P100-C60-G50  & 73.82\%     & F1-P400-C5-G50   & 71.01\%     & F30-P300-C100-G50  & 82.30\%     \\
F1-P150-C5-G50   & 71.69\%     & F1-P450-C5-G50   & 72.59\%     & F40-P300-C100-G50  & 83.88\%     \\
F1-P200-C120-G50 & 75.01\%     & F1-P50-C25-G50   & 68.31\%     & F40-P300-C100-G50  & 83.88\%     \\
F1-P200-C5-G50   & 74.47\%     & F1-P50-C30-G50   & 66.40\%     &                    &                \\
\bottomrule
\end{tabular}
\end{table*}

\begin{table*}[h]
\caption{Most frequent selected features and their presence in some experiments.}
\label{tbl:featurefrequencyinexp}
\centering
\setlength{\tabcolsep}{11pt}
\begin{tabular}{lcccccc}
\toprule
\multicolumn{3}{r}{ExpID:} & F10-P400-C240-G50 & F20-P300-C100-G50 & F1-P400-C240-G50 & F1-P300-C180-G50 \\
\multicolumn{3}{r}{BestFitness:} & 84.85\% & 83.99\% & 80.99\% & 77.92\% \\
\midrule
FeatureText & FNo & Freq. &  &  &  &  \\
\midrule
co & 2954 & 32 & \cellcolor{gray!25}TRUE & \cellcolor{gray!25}TRUE & \cellcolor{gray!50}FALSE & \cellcolor{gray!25}TRUE \\
free & 4885 & 30 & \cellcolor{gray!25}TRUE & \cellcolor{gray!25}TRUE & \cellcolor{gray!25}TRUE & \cellcolor{gray!25}TRUE \\
enter & 4264 & 28 & \cellcolor{gray!25}TRUE & \cellcolor{gray!25}TRUE & \cellcolor{gray!25}TRUE & \cellcolor{gray!25}TRUE \\
shelv & 11102 & 27 & \cellcolor{gray!50}FALSE & \cellcolor{gray!50}FALSE & \cellcolor{gray!50}FALSE & \cellcolor{gray!25}TRUE \\
http & 5920 & 26 & \cellcolor{gray!25}TRUE & \cellcolor{gray!50}FALSE & \cellcolor{gray!25}TRUE & \cellcolor{gray!25}TRUE \\
porn & 9685 & 26 & \cellcolor{gray!25}TRUE & \cellcolor{gray!25}TRUE & \cellcolor{gray!25}TRUE & \cellcolor{gray!25}TRUE \\
election2juli & 4177 & 25 & \cellcolor{gray!50}FALSE & \cellcolor{gray!25}TRUE & \cellcolor{gray!50}FALSE & \cellcolor{gray!50}FALSE \\
foogo & 4794 & 25 & \cellcolor{gray!25}TRUE & \cellcolor{gray!50}FALSE & \cellcolor{gray!50}FALSE & \cellcolor{gray!50}FALSE \\
juju & 6690 & 23 & \cellcolor{gray!50}FALSE & \cellcolor{gray!25}TRUE & \cellcolor{gray!50}FALSE & \cellcolor{gray!50}FALSE \\
carr & 2571 & 22 & \cellcolor{gray!50}FALSE & \cellcolor{gray!25}TRUE & \cellcolor{gray!50}FALSE & \cellcolor{gray!50}FALSE \\
elimin & 4190 & 22 & \cellcolor{gray!50}FALSE & \cellcolor{gray!25}TRUE & \cellcolor{gray!50}FALSE & \cellcolor{gray!50}FALSE \\
lifetim & 7304 & 22 & \cellcolor{gray!25}TRUE & \cellcolor{gray!50}FALSE & \cellcolor{gray!50}FALSE & \cellcolor{gray!50}FALSE \\
mybodym & 8300 & 22 & \cellcolor{gray!50}FALSE & \cellcolor{gray!50}FALSE & \cellcolor{gray!50}FALSE & \cellcolor{gray!25}TRUE \\
overdose-revers & 9194 & 22 & \cellcolor{gray!50}FALSE & \cellcolor{gray!50}FALSE & \cellcolor{gray!50}FALSE & \cellcolor{gray!25}TRUE \\
rqanroxc47 & 10618 & 22 & \cellcolor{gray!50}FALSE & \cellcolor{gray!25}TRUE & \cellcolor{gray!25}TRUE & \cellcolor{gray!25}TRUE \\
sabha & 10713 & 22 & \cellcolor{gray!50}FALSE & \cellcolor{gray!50}FALSE & \cellcolor{gray!50}FALSE & \cellcolor{gray!50}FALSE \\
wbdwyyxbgj & 13620 & 22 & \cellcolor{gray!50}FALSE & \cellcolor{gray!25}TRUE & \cellcolor{gray!50}FALSE & \cellcolor{gray!50}FALSE \\
xboxon & 13976 & 22 & \cellcolor{gray!50}FALSE & \cellcolor{gray!25}TRUE & \cellcolor{gray!50}FALSE & \cellcolor{gray!25}TRUE \\
clock & 2910 & 21 & \cellcolor{gray!50}FALSE & \cellcolor{gray!25}TRUE & \cellcolor{gray!50}FALSE & \cellcolor{gray!50}FALSE \\
continu & 3131 & 21 & \cellcolor{gray!50}FALSE & \cellcolor{gray!50}FALSE & \cellcolor{gray!50}FALSE & \cellcolor{gray!50}FALSE \\
darrellejon & 3479 & 21 & \cellcolor{gray!25}TRUE & \cellcolor{gray!50}FALSE & \cellcolor{gray!50}FALSE & \cellcolor{gray!50}FALSE \\
gtukhvetov4 & 5410 & 21 & \cellcolor{gray!50}FALSE & \cellcolor{gray!50}FALSE & \cellcolor{gray!50}FALSE & \cellcolor{gray!50}FALSE \\
ibac & 6021 & 21 & \cellcolor{gray!50}FALSE & \cellcolor{gray!50}FALSE & \cellcolor{gray!50}FALSE & \cellcolor{gray!50}FALSE \\
luafkcop2r & 7524 & 21 & \cellcolor{gray!50}FALSE & \cellcolor{gray!50}FALSE & \cellcolor{gray!50}FALSE & \cellcolor{gray!50}FALSE \\
reachabl & 10226 & 21 & \cellcolor{gray!50}FALSE & \cellcolor{gray!50}FALSE & \cellcolor{gray!50}FALSE & \cellcolor{gray!50}FALSE \\
sportscent & 11552 & 21 & \cellcolor{gray!50}FALSE & \cellcolor{gray!50}FALSE & \cellcolor{gray!25}TRUE & \cellcolor{gray!50}FALSE \\
ude29 & 13014 & 21 & \cellcolor{gray!50}FALSE & \cellcolor{gray!50}FALSE & \cellcolor{gray!50}FALSE & \cellcolor{gray!50}FALSE \\
z55nuvynrw & 14229 & 21 & \cellcolor{gray!50}FALSE & \cellcolor{gray!50}FALSE & \cellcolor{gray!50}FALSE & \cellcolor{gray!50}FALSE \\
zixqiz7fjq & 14285 & 21 & \cellcolor{gray!25}TRUE & \cellcolor{gray!25}TRUE & \cellcolor{gray!25}TRUE & \cellcolor{gray!25}TRUE \\
\bottomrule
\end{tabular}
\end{table*}

\subsection{Model Validation}
The selected features and the optimized parameters (i.e.,, the configurations that attained high fitness values) in some experiments are used to model spam prediction using XGBoost. The model robustness is validated using 10-Fold cross-validation repeated 50 times. Table \ref{tbl:comparisionGACrossVal50} summarizes the absolute difference between the GA fitness value and the validated XGBoost model performance. 

\begin{table}[H]   
\caption{Comparison between selected best fitness obtained by GA optimization and the results of 10-fold cross-validation repeated 50 times.}\label{tbl:comparisionGACrossVal50}
\begin{tabular}{p{0.34\linewidth} p{0.1\linewidth}  p{0.1\linewidth} p{0.10\linewidth} p{0.09\linewidth}}
\toprule
ExpID   & Feat.   & GA Fitness & GMean Avg. (SD) & Diff. \\
\midrule
F1-P300-C180-G50    & 140 & 77.92           & 75.10 (0.033) &   2.82    \\
F1-P400-C240-G50  & 141 & 80.99           & 78.56 (0.031) &    \textbf{2.43} \\
F20-P300-C100-G50 & 2638 &   83.99           & 80.60 (0.032)   &  3.39 \\
F10-P400-C240-G50 & 1355 &\textbf{84.85}           & \textbf{82.32} (\textbf{0.030})   &  2.53 \\

\bottomrule
\end{tabular}
\end{table}

The best optimization and feature selection experiment ``F10-P400-C240-G50'' attained a fitness value that equals 84.85\% (i.e., Geometric Mean). The outcomes presented in Table \ref{tbl:OptimizedParams} are used to build the spam prediction model that is validated using 10-fold stratified cross-validation and repeated 50 times.

\begin{table}[H]
\caption{Optimized XGBoost parameters and the number of selected features obtained by the experiment ``F10-P400-C240-G50''.}
\label{tbl:OptimizedParams}
\centering
\begin{tabular}{lc}
\hline
XGBoost Parameters    & Optimized Value \\
\hline
learning\_rate        & 0.47            \\
n\_estimators         & 93              \\
max\_depth            & 4               \\
min\_child\_weight    & 0.08            \\
gamma                 & 0.42            \\
subsample             & 0.94            \\
colsample\_bytree     & 0.84            \\
\hline
\textit{Number of Selected Features} & \textit{1355}           \\
\hline
\end{tabular}
\end{table}

\begin{figure}[H]
	\centering
		\includegraphics[scale=0.6]{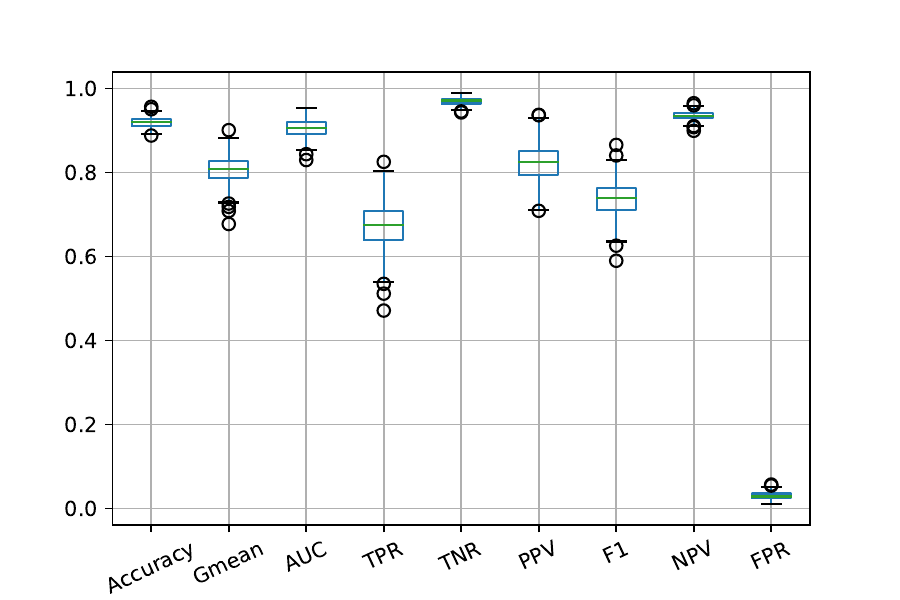}
	\caption{Results of experiment F10-P400-C240-G50.}
	\label{FIG:50RunsBoxP}
\end{figure}

\begin{figure*}[h]
	\centering
		\includegraphics[scale=0.6]{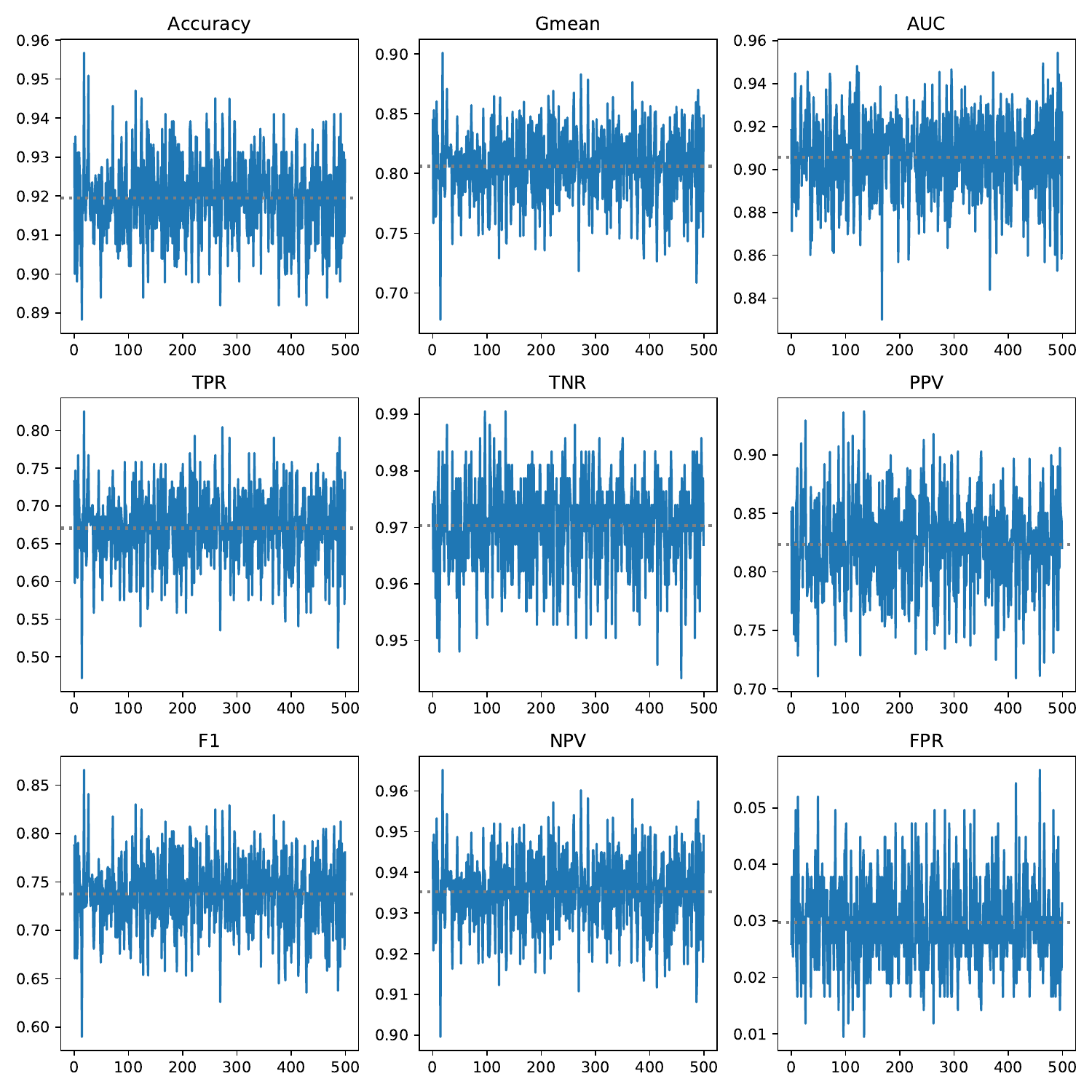}
	\caption{Results of experiment F10-P400-C240-G50.}
	\label{FIG:50RunsLineChart}
\end{figure*}

XGBoost has been initialized with the optimized parameters in Table \ref{tbl:OptimizedParams}, and used the dataset with the selected features to build the spam classification model. The run is validated using 10-fold stratified cross-validation and repeated 50 times. Figure \ref{FIG:50RunsBoxP} is a box plot that summarizes the performance metrics, and Figure \ref{FIG:50RunsLineChart} depicts the performance metrics of each fold. The detailed performance metrics of the 10x50CV are presented in Table \ref{tbl:resultsF10-P400-C240-G50}.

\begin{table}[H]  
\caption{Results of experiment ``F10-P400-C240-G50''  repeated 50 times with 10-Fold cross-validation per each run. (Best fitness obtained by GA was GMean = 84.85\%)}\label{tbl:resultsF10-P400-C240-G50}
\centering
\setlength{\tabcolsep}{14pt}
\begin{tabular}{lcccc}
\toprule
Metric & Min.  & Avg.  & Max.  & SD    \\
\midrule
Accuracy             & 88.82 & 92.67 & 95.88 & $\pm0.010$ \\
GMean                & 72.71 & 82.32 & 91.26 & $\pm0.030$ \\
AUC                  & 86.54 & 92.72 & 97.29 & $\pm0.018$ \\
TPR                  & 54.02 & 69.68 & 84.88 & $\pm0.050$ \\
TNR                  & 94.8  & 97.37 & 99.29 & $\pm0.008$ \\
PPV                  & 70.27 & 84.54 & 95.59 & $\pm0.039$ \\
FPR                  & 0.71  & 2.63  & 5.2   & $\pm0.008$ \\
F1                   & 64.6  & 76.28 & 87.43 & $\pm0.037$ \\
NPV                  & 91.19 & 94.02 & 96.97 & $\pm0.009$
\\
\bottomrule
\end{tabular}
\end{table}

There are two main observations worth mentioning here (a) the GA feature selection and parameter optimization effectiveness are noticeably acceptable, and (b) XGBoost classification model is barely affected by the algorithm randomness. The absolute difference between the GA fitness and the validated model is bound to an average of approximately 2.8, and the standard deviation of all the evaluation measures is less than 0.04 after 50 cross-validation runs.

\subsection{Statistical Analysis}

The effect of the stochastic nature of GA, and the imposed randomness of the classification algorithms on the experiments are described using the statistical tests. The whole process of GA-based parameter optimization and feature selection followed by a 10x50CV of XGBoost is repeated seven times. Runs denoted by the sequence ``R01'' to ``R07'' follow exactly the same steps as the run ``F10-P400-C240-G50''; and to ensure non restricted random number generation the seed is not fixed to a certain value in any step. The run ``F10-P400-C240-G50'' is the main baseline for the comparison and denoted by ``R00''. Table \ref{tbl:descStatRuns} lists the descriptive statistics of the GMean values of the runs. Only one of the runs ``R04'' showed a slight drift in the GMean value less than 2\% below the average of all runs.

\begin{table}
\caption{Descriptive statistics of GMean values of different runs. (8 runs of ``F10-P400-C240-G50'' experiment, each is validated 50x10CV)}\label{tbl:descStatRuns}
\centering
\setlength{\tabcolsep}{4.5pt}
\begin{tabular}{lccccccc}
\hline
Run & Mean \% & $\pm$ SD & Min.  \% & 25  \%  & 50  \% & 75  \% & Max  \% \\
\hline
R00 & 82.32 & 0.030 & 72.71 & 80.42 & 82.53 & 84.44 & 91.26 \\
R01 & 81.48 & 0.031 & 67.83 & 79.58 & 81.47 & 83.50 & 92.10 \\
R02 & 81.46 & 0.030 & 68.96 & 79.74 & 81.45 & 83.38 & 90.00 \\
R03 & 81.57 & 0.028 & 72.02 & 79.66 & 81.65 & 83.47 & 90.19 \\
R04 & 78.34 & 0.031 & 64.24 & 76.35 & 78.53 & 80.25 & 88.85 \\
R05 & 81.10 & 0.030 & 70.98 & 79.43 & 81.21 & 83.02 & 89.89 \\
R06 & 81.14 & 0.030 & 73.71 & 79.30 & 81.21 & 83.04 & 92.05 \\
R07 & 82.46 & 0.029 & 72.35 & 80.64 & 82.26 & 84.35 & 90.40 \\
\hline
\textbf{Mean}& \textbf{81.23} & \textbf{0.0300}&	\textbf{70.35}&	\textbf{79.39}&	\textbf{81.29}&	\textbf{83.18}&	\textbf{90.59} \\
\hline
\end{tabular}
\end{table}

Further nonparametric statistical tests \cite{brownlee2018statistical,whitney1971direct,WoolsonWilcoxon,chan1997learning} illustrate the level of similarity between the performance metrics of the runs. The tests analyse the GMean values of each run folds and the other runs. Namely, Wilcoxon Signed-Rank Test \cite{WoolsonWilcoxon,brownlee2018statistical,whitney1971direct} between the run pairs and Kruskal statistical test \cite{chan1997learning,brownlee2018statistical,whitney1971direct} of run combinations.

The p-value of Wilcoxon statistical test between run pairs in Table \ref{tbl:WilcoxonTest} indicates a possible similarity between the runs ``R00'' and ``R07''; considering $\alpha = 0.05$. The differences against the rest of the runs are possibly because the run ``R00'' is controlled in terms of a random number generation method. According to the Wilcoxon test, the runs ``R01, R02, R03, R05, and R06'' possibly have similar distributions, ``R04'' distribution is different from the others. However, the GMean value of the runs does not significantly differ as illustrated in Table \ref{tbl:descStatRuns}.

\begin{table}[H]
\caption{p-Value of Wilcoxon statistical test between run pairs}\label{tbl:WilcoxonTest}
\centering
\setlength{\tabcolsep}{4.2pt}
\def\arraystretch{1.2}
\begin{tabular}{cccccccc}
\hline
\textbf{R01} & 0.0001 &  &  &  &  &  &   \\
\textbf{R02} & 0.00002 & 0.76166 &  &  &  &  &   \\
\textbf{R03} & 0.00013 & 0.56297 & 0.51467 &  &  &  &  \\
\textbf{R04} & 0 & 0 & 0 & 0 &  &  &   \\
\textbf{R05} & 0 & 0.05551 & 0.0756 & 0.03146 & 0 &  &  \\
\textbf{R06} & 0 & 0.05079 & 0.05912 & 0.01797 & 0 & 0.98696 &  \\
\textbf{R07} & 0.70065 & 0 & 0 & 0 & 0 & 0 & 0  \\
    & \textbf{R00} & \textbf{R01} & \textbf{R02} & \textbf{R03} & \textbf{R04} & \textbf{R05} & \textbf{R06} \\
\hline
\end{tabular}
\end{table}

Kruskal statistical test applies to three or more run combinations. Therefore, all run combinations are tested and the top p-values of the nonparametric Kruskal test are reported in Table \ref{tbl:KruskalTestCombinations}. Assuming an $\alpha = 0.05$, the runs ``R01, R02, and R05'' are expected to have the highest similarity of GMean distributions, and ``R01, R02, R03, R05, and R06'' most probably have  similar distributions.

\begin{table}[H]
\caption{Kruskal statistical test of run combinations, showing only top combinations.}\label{tbl:KruskalTestCombinations}
\centering
\begin{tabular}{lc}
\hline
Run & p-Value \\
\hline
R01, R02, R05 & 0.14449 \\
R01, R02, R03, R05 & 0.14072 \\
R02, R05, R06 & 0.13411 \\
R01, R05, R06 & 0.09977 \\
R01, R02, R06 & 0.09401 \\
R01, R02, R05, R06 & 0.09125 \\
R01, R02, R03, R06 & 0.08424 \\
R02, R03, R05 & 0.07825 \\
R01, R03, R05 & 0.07208 \\
R01, R02, R03, R05, R06 & 0.04973 \\
R02, R03, R06 & 0.04358 \\
R01, R03, R06 & 0.04145 \\
R02, R03, R05, R06 & 0.03897 \\
R01, R03, R05, R06 & 0.03371 \\
R03, R05, R06 & 0.0312 \\
\hline
\end{tabular}
\end{table}

\subsection{Comparison Analysis}

\subsubsection{Comparison with related work}
The overall performance and robustness of the proposed spam classification model outperforms the work in \cite{jain2019optimizing}. The authors in \cite{jain2019optimizing} claim a high performing LSTM (Long Short Term Memory) model, however the robustness of their approach is not well justified. The performance metrics are based on the ``Ham'' class as the positive class which is the majority class label. Swapping minority class with majority class will lead to significantly higher performance metrics values; which is misleading when interpreting some metrics such as TPR. There is no evidence of using cross-validation in assessing the model robustness. Features are reduced by the selection of most frequent words in the corpus, and the configuration of the basic classifiers used in performance comparison is not presented in the paper. 
For the sake of fair comparison with the results of our research, the performance metrics are re-calculated to consider ``Ham'' as the positive class label. Table \ref{tbl:metricsConversion} shows the equivalence equations used to re-calculate the results of our experiments to be comparable with the results in \cite{jain2019optimizing}. Nonetheless, Table \ref{tbl:Ref1Metrics} compares the performance metrics of \cite{jain2019optimizing} in spam classification (having ``Ham'' as positive class, and ``Spam'' as negative class) and Table \ref{tbl:Ref1FeatSelAcc} confronts the effect of feature reduction.

\begin{table}[H]
\caption{Metrics re-calculation equations based on the selection of the positive class, for the sake of comparison. \textit{In our research P denotes ``Spam'', and N denotes ``Ham''}}\label{tbl:metricsConversion}
\centering
\setlength{\tabcolsep}{4pt}
\renewcommand{\arraystretch}{2.5}
\begin{tabular}{llll}
\toprule
Metric & \thead{Positive\\ ``Spam''} & \thead{Positive\\ ``Ham''} & \thead{Equivalence\\in our work} \\
\midrule
Recall & $\frac{TP}{TP+FN}$ & $\frac{TN}{TN+FP}$ & $1 - FPR$ \\

Precision &  $\frac{TP}{TP+FP}$ & $\frac{TN}{TN+FN}$ & $NPV$ \\ 

F1 & $2 \times \frac{TPR \times PPV}{TPR + PPV}$ & $2 \times \frac{TNR \times NPV}{TNR + NPV}$ & $2 \times \frac{TNR \times NPV}{TNR + NPV}$ \\
\bottomrule
\end{tabular}
\end{table}




\begin{table}[H]  
\caption{Comparison of the performance metrics in \cite{jain2019optimizing} with the best results of our experiments after re-calculating the performance metrics (i.e., considering ``Ham'' as the positive class).}
\label{tbl:Ref1Metrics}
\centering
\setlength{\tabcolsep}{10pt}
\begin{tabular}{lllll}
\toprule
Classifier & Precision & Recall & Accuracy & F1    \\
\midrule
KNN        & 91.61     & 91.96  & 91.96    & 91.38 \\
NB         & 91.69     & 92.06  & 92.06    & 91.74 \\
RF         & 93.25     & 93.43  & 93.43    & 93.04 \\
ANN        & 91.80     & 91.18  & 91.18    & 91.41 \\
SVM        & 92.91     & 93.14  & 93.14    & 92.97 \\
SLSTM      & 95.54     & 98.37  & 95.09    & 96.84  \\
\midrule
\multicolumn{5}{l}{\textit{*Experiment F10-P400-C240-G50, $GMean=82.32\pm0.030$}} \\
Avg+      &   94.02   & 97.37  &  92.67  &  95.67\\
Min+      &   91.19   & 94.8 &  88.82   &  \\ 
Max+     &   96.97   & 99.29  &  95.88   & \\ 
SD      &   $\pm0.009$   & $\pm0.008$  &   $\pm0.010$  &   \\
\bottomrule
\multicolumn{5}{l}{\makecell[l]{*$50\times10 CV$ \\+Recalculated to consider ``Ham'' as positive class\\ according to the definitions in Table \ref{tbl:metricsConversion} }}
\end{tabular}
\end{table}

\begin{table}[H]  
\caption{Effect of feature reduction in \cite{jain2019optimizing} compared to some experiments in our research.}
\label{tbl:Ref1FeatSelAcc}
\centering
\begin{tabular}{lll}
\toprule
\multicolumn{3}{l}{\textit{Frequency-based feature reduction in \cite{jain2019optimizing}}}\\
Features     & Accuracy \% & Notes\\
\midrule
5000   & 93.81 &                         \\
8000   & 93.91   &                       \\
10,000 & 94.21   &                       \\
14,000 & 95.09   &   \\
\midrule
\multicolumn{3}{l}{\textit{Some experiments in our research ($50\times10  CV)$}} \\
Features & Accuracy\% $\pm SD$ \textit{(maximum)} & Experiment ID \\
\midrule
1355 & $92.67\pm0.010$ \textit{(95.88)} & F10-P400-C240-G50\\
2638  &$91.94\pm0.011$ \textit{(95.68)} & F20-P300-C100-G50 \\
141 & $91.66\pm0.010$  \textit{(94.71)}& F1-P400-C240-G50 \\
140 & $90.61\pm0.010$  \textit{(93.53)}& F1-P300-C180-G50 \\
142 & $88.57\pm0.011$  \textit{(91.75)}& F1-P100-C60-G50 \\
144 & $84.95\pm0.012$  \textit{(88.24)}& F1-P10-C6-G100 \\  

\bottomrule
\end{tabular}
\end{table}

The modified GA approach in this study outperforms the approach used in \cite{jain2019optimizing} in terms of feature selection, The number of features selected in the proposed approach (i.e., 1355 features) is much lower than the number of features selected  in \cite{jain2019optimizing} (i.e.,, 5000 features). On the other hand, the maximum accuracy obtained using GA in our approach was 95.88\% compared to 95.09\% in \cite{jain2019optimizing}.

\subsubsection{Comparison with $Chi^2$ Feature Selection}

$Chi^2$ \cite{cekik2022new,liu1995chi2} statistical test has been used in text feature selection based on statistical significance of features. We selected the top ``1355 features'' using the $Chi^2$ method to compare the results with our best findings. Different machine learning algorithms are validated by 10x50CV in building spam prediction models using ``1355'' selected features by $Chi^2$. The algorithms used without parameter tuning are XGBoost, Multinomial Naive Bayes (MNNB), K-Nearest Neighbors (KNN), Logistic Regression (LR), Adaptive Boosting (AdaBoost), and Decision Trees (DT). The p-value of Wilcoxon statistical test between the run ``F10-P400-C240-G50'' and $Chi^2$-based models are presented in Table \ref{tbl:WilcoxonChi2}, and the descriptive statistics of the performance metrics are shown in Table \ref{tbl:MetricsCompChi2}.

\begin{table}[H]
\caption{p-Value of Wilcoxon statistical comparing GMean of ``F10-P400-C240-G50'' with $Chi^2$ feature selection method performance.}\label{tbl:WilcoxonChi2}
\centering
\begin{tabular}{lc}
\hline
Experiment & p-Value \\
\hline
XGBoost \& $Chi^2$ & 0.00000 \\
MNNB  \& $Chi^2$ & 0.00000 \\
KNN  \& $Chi^2$ & 0.00000 \\
LR  \& $Chi^2$& 0.00000 \\
AdaBoost \& $Chi^2$ & 0.00000 \\
DT  \& $Chi^2$ & 0.51423 \\
\hline
\end{tabular}
\end{table}

\begin{table*}
\caption{Comparison between the run 'F10-P400-C240-G50' with $Chi^2$ feature selection method performance. (each is validated by 50x10CV)}\label{tbl:MetricsCompChi2}
\centering
\begin{tabular}{lccccccc}
\hline
Metric & F10-P400-C240-G50 & XGBoost \& $Chi^2$   & MNNB \& $Chi^2$    & KNN3 \& $Chi^2$    & LR \& $Chi^2$    & AdaBoost \& $Chi^2$    & DT \& $Chi^2$   \\
\hline
Accuracy & \textbf{92.67} (0.010) & 92.06 (0.009) & 88.77 (0.009) & 87.17 (0.009) & 91.34 (0.009) & 92.46 (0.010) & 91.38 (0.011) \\
GMean & \textbf{82.32} (0.030) & 76.12 (0.032) & 58.13 (0.044) & 50.88 (0.049) & 71.46 (0.037) & 80.60 (0.030) & 82.20 (0.028) \\
AUC & 92.72 (0.018)& 93.79 (0.016) & \textbf{95.61} (0.012) & 77.13 (0.029) & 95.39 (0.012) & 92.82 (0.018) & 83.31 (0.024) \\
TPR & 69.68 (0.050) & 58.71 (0.049) & 33.99 (0.051) & 26.23 (0.050) & 51.46 (0.054) & 66.55 (0.049) & \textbf{70.77} (0.047) \\
TNR & 97.37 (0.008) & 98.88 (0.005) & \textbf{99.96} (0.001) & 99.63 (0.003) & 99.50 (0.003) & 97.76 (0.007) & 95.59 (0.010) \\
PPV & 84.54 (0.039) & 91.58 (0.035) & \textbf{99.49} (0.012) & 93.61 (0.051) & 95.51 (0.028) & 86.00 (0.041) & 76.81 (0.042) \\
FPR & 2.63 (0.008) & 1.12 (0.005) & \textbf{0.04} (0.001) & 0.37 (0.003) & 0.50 (0.003) & 2.24 (0.007) & 4.41 (0.010) \\
F1s & \textbf{76.28} (0.037) & 71.42 (0.039) & 50.46 (0.057) & 40.73 (0.063) & 66.71 (0.047) & 74.9 (0.038) & 73.56 (0.035) \\
NPV & 94.02 (0.009) & 92.14 (0.009) & 88.11 (0.008) & 86.86 (0.008) & 90.94 (0.009) & 93.47 (0.009) & \textbf{94.12} (0.009) \\
\hline
\end{tabular}
\end{table*}

The p-value of the Wilcoxon test indicates different distributions of GMean value compared to the run ``F10-P400-C240-G50''. except for the DT model. The similar distributions are justified by the fact that XGBoost is an evolution of the decision tree algorithm; they share similar characteristics that could lead to similar behavior. However, ``F10-P400-C240-G50'' model outperforms the DT model as indicated by the majority of the performance metrics.

\subsubsection{Comparison with PCA Feature Selection}

\begin{figure}[H]
	\centering
		\includegraphics[scale=0.6]{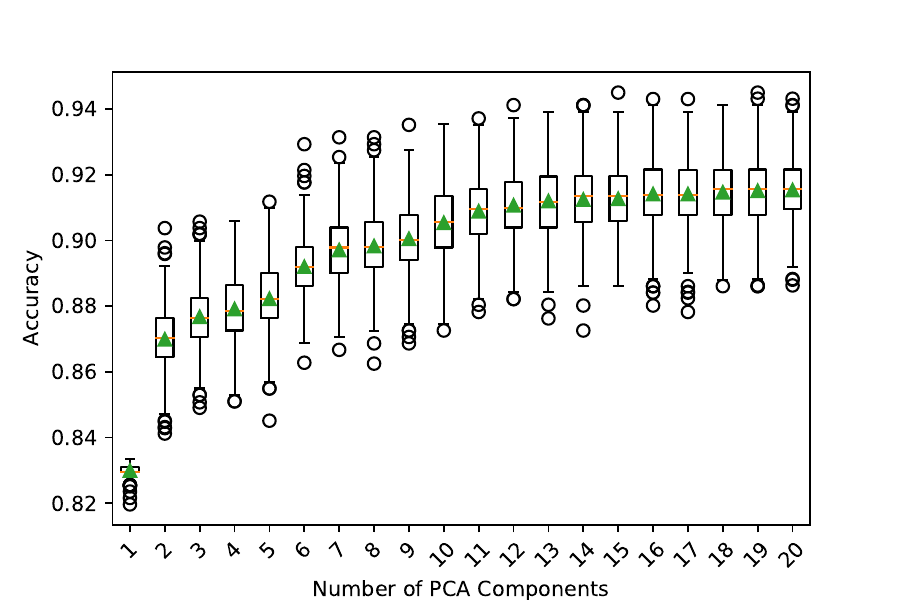}
	\caption{PCA Feature Selection and XGBoost. (Validated 50x10CV)}
	\label{FIG:PCAXGB-BoxP}
\end{figure}

Reducing the dimension of relatively large feature space while preserving most of the information is possible using Principal Component Analysis (PCA) \cite{jolliffe2005principal}. The feature set is transformed into a number of principal components based on their covariance matrix, then a relatively small number of the principal components will be selected to represent the full feature set. We trained an XGBoost model using a different number of principal components (i.e., from 1 to 20) and validated the model using 50x10CV. An illustration of the model accuracy in relation to the number of principal components is shown in Figure \ref{FIG:PCAXGB-BoxP}. The accuracy metrics and the standard deviation are presented in Table \ref{tbl:MetricsPCA-XGB}.

\begin{table}[H]
\caption{Accuracy of PCA feature selection and XGBoost. (each is validated by 50x10CV)}\label{tbl:MetricsPCA-XGB}
\centering
\begin{tabular}{ccc}
\hline
PCA Components & Accuracy \% & SD $\pm$\\
\hline
01  & 82.99 & 0.001 \\
02  & 86.98 & 0.009 \\
03  & 87.68 & 0.009 \\
04  & 87.91 & 0.010 \\
05  & 88.22 & 0.010 \\
06  & 89.20 & 0.010 \\
07  & 89.70 & 0.010 \\
08  & 89.82 & 0.010 \\
09  & 90.04 & 0.011 \\
10 & 90.53 & 0.011 \\
11 & 90.88 & 0.011 \\
12 & 91.07 & 0.010 \\
13 & 91.20 & 0.011 \\
14 & 91.24 & 0.011 \\
15 & 91.27 & 0.010 \\
16 & 91.41 & 0.010 \\
17 & 91.41 & 0.010 \\
18 & 91.46 & 0.011 \\
19 & 91.51 & 0.010 \\
20 & 91.53 & 0.010 \\
\hline
\end{tabular}
\end{table}

It is apparent that 20 principal components will enable attaining 91.53\% total accuracy in spam prediction; compared to 92.67\% using our modified GA. Moreover, PCA will reduce significantly the feature space but makes the model interpretation much harder. It is apparent that the PCA based model will converge after 15 PCA components. According to the illustration in Figure \ref{FIG:PCAXGB-BoxP} and Table \ref{tbl:MetricsPCA-XGB} the improvement in accuracy was less than 0.5 percentage absolute point within the last nine components (i.e., PCA components 12-20).

Therefore, PCA-based XGBoost models under-perform the modified GA-based XGboost models.

\subsubsection{Comparison with BERT and Deep Learning}

Most recent advancements in natural language processing research introduced pre-trained word embedding models that are coupled with Deep Learning (DL) algorithms \cite{borse2022state,taneja2022comparison,oswald2022spotspam}. BERT word embedding is used with DL to build spam classification models. The major issue of interest in DL is the computational complexity and extensive resource use. Despite such limitations, we were able to build a spam prediction model in this research and validated with the percentage split of the tweets dataset. BERT is used in text pre-processing and encoding, ``sigmoid'' activation function, Tensorflow \cite{pang2020deep}, and Keras \cite{ketkar2017introduction}. 
Table \ref{tbl:MetricsBERT-DL} summarizes the major performance metrics of the generated model over 20 epochs. The TPR of the class of interest (``Spam'') is relatively low (52\%) which indicates a very low prediction power of the generated model in spam prediction. 

\begin{table}[H]
\caption{Accuracy of BERT model and DL. (20 epochs)}\label{tbl:MetricsBERT-DL}
\centering
\begin{tabular}{lcccc}
\hline
class & PPV (precision)  &  TPR (recall) & f1  & support\\
\hline
'ham'    &   0.91   &   0.97    &  0.94   &   1058 \\
'spam'   &    0.79  &    0.52   &   0.63   &    216 \\

\textit{accuracy}  &     &     & \textbf{0.90} &     1274\\
\textit{macro avg}   &    0.85   &   0.75    &  0.78  &   1274\\
\textit{weighted avg}   & 0.89   &   0.90  &    0.89   &   1274\\
\hline
\end{tabular}
\end{table}

For the sake of comparison with recent advancement in text classification we implemented the BERT-DL model to assess its feasibility in spam prediction. In our case, the limited computational resources were the main barrier in tuning and seeking better prediction performance. However, the experiment shows that our modified GA approach outperforms the DL approach. Moreover, the resulting models are not intuitive to be interpreted and starve for computational resources. 

\subsubsection{Experimenting with SMS Dataset}

The modified GA is applied to a public imbalanced SMS dataset\cite{almeida2011contributions}; about 13\% of the messages are ``Spam''. Hyper parameter optimization and feature selection results are listed in Table \ref{tbl:SMSOptimizedParams}. The GA reduced the selected features to 9.52\% of total dataset features (i.e., 706 out of 7419 features) attained a GMean value of 97.29\%. 

\begin{table}[H]
\caption{Optimized XGBoost parameters and the number of selected features obtained by the ``SMS Spam Dataset'' experiment using modified GA and XGBoost.}
\label{tbl:SMSOptimizedParams}
\centering
\begin{tabular}{lc}
\hline
XGBoost Parameters    & Optimized Value \\
\hline
learning\_rate        & 0.31            \\
n\_estimators         & 47              \\
max\_depth            & 10               \\
min\_child\_weight    & 0.15            \\
gamma                 & 0.33            \\
subsample             & 1.0            \\
colsample\_bytree     & 0.69            \\
\hline
\textit{Number of Selected Features} & \textit{706}           \\
\hline
\end{tabular}
\end{table}

The outcomes in Table \ref{tbl:SMSOptimizedParams} initialized an XGBoost classifier to model SMS spam. The model is validated using a 50 times repeated run of 10-fold stratified cross-validation. Table \ref{tbl:SMSXGBsummary} shows the performance metrics.

\begin{table}[H]  
\caption{Results of ``SMS Spam Dataset'' using modified GA and XGBoost; repeated 50 times with 10-Fold cross-validation per each run. (Best fitness obtained by GA was GMean = 97.29\%)}\label{tbl:SMSXGBsummary}
\centering
\setlength{\tabcolsep}{14pt}
\begin{tabular}{lcccc}
\toprule
Metric & Min.  & Avg.  & Max.  & SD    \\
\midrule
Accuracy & 95.15 & 97.4  & 99.1  & 0.006 \\
GMean    & 85.94 & 92.49 & 98.35 & 0.021 \\
AUC      & 93.08 & 97.99 & 99.85 & 0.011 \\
TPR      & 74.32 & 86.35 & 97.33 & 0.039 \\
TNR      & 97.72 & 99.12 & 100   & 0.004 \\
PPV      & 84.29 & 93.88 & 100   & 0.028 \\
FPR      & 0     & 0.88  & 2.28  & 0.004 \\
F1s      & 81.38 & 89.89 & 96.69 & 0.025 \\
NPV      & 96.18 & 97.92 & 99.59 & 0.006 \\
\bottomrule
\end{tabular}
\end{table}


In comparison to the best results in \cite{abid2022spam,https://doi.org/10.1002/cpe.6989,9777157} the modified GA shows a competitive performance.  The authors in \cite{abid2022spam}, \cite{https://doi.org/10.1002/cpe.6989} ,and \cite{9777157} attained maximum of total accuracy equals 96\%, 96.8\% and 98.74\% respectively. Its worth mentioning that Random Forest and SVM algorithms attained 99\% accuracy in\cite{abid2022spam} but with TF-iDF features and oversampling. In essence, the maximum accuracy attained by the modified GA was 99.1\%; which makes it outperform the majority of the related works utilizing the same SMS dataset. 

Our proposed approach, modified GA, reduces all the features of the tweets by 9.45\% (i.e., from 14343 to 1355 features) and maintains a competitive performance in comparison to the related studies. Therefore, the proposed approach in this research is expected to reduce the dimensionality  by automating the process of feature selection and tuning the prediction model parameters simultaneously. The results of this work could be extended to list the features as words (i.e., specific words in the tweets) for further feature analysis. Furthermore, the XGBoost tree models have a higher level of interpretability compared to ANN-based models; which make it much easier to deeply analyse the models for the sake of spam understanding and modeling.

\subsection{Implications and Limitations}

The reported experiments and outcomes of this research establish a basis for spam modeling. In essence, it outperformed many related works in SMS spam modeling. Future research may build on the outcomes to enhance understanding of spam behavior. Further, this research could be considered for generalization in other domains such as software engineering, construction engineering, internet of things, and smart cities. In contrast to black-box models, tree-based classifiers enable straightforward implementation to detect spam tweets. The tremendous growth of Online Social Networks (OSN) calls for efficient real-time spam detectors. The state of the art solutions recommend Deep Learning based solutions, however deep learning is resource consuming and overlooks unseen spam behaviors. Our proposed approach reduces learning time significantly compared to deep learning based solutions. 

Usually, GA finds outstanding solutions once its parameters (i.e., Initial population, mutation and cross-over ratios, number of generations, ...) are well tuned. In this research tuning the GA using Grid Search is time consuming. Therefore, sensitivity analysis described in section \ref{sub:SA} has been used to find the best GA parameters. In the near future we expect the reliability of the public twitter spam dataset to raise concerns due to subjective interpretations by different communities. Multi class labeling of spam text in sentiment analysis is not considered in this research.

The large number of experiment runs and the comprehensive set of performance metrics would direct further research activities.

\section{Conclusions and Future Directions}
\label{Sec:Con}
 Spam modeling is a challenging task due to many issues such as the high dimensionality of the features space, the imbalanced class distributions, the bias of classification algorithms towards the majority class, and natural language processing issues. Many of the related research works lack solid validation of the generated models and usually report positive class-based performance metrics. In this paper, a modified genetic algorithm is designed in order to perform two main tasks; (1) an effective dimensionality reduction of an imbalanced tweets dataset and (2) hyper parameter optimization of XGBoost classification algorithm. Intensive validation of the generated prediction model illustrates the robustness of the modified algorithm and its competitive performance compared to other approaches. This research reports a comprehensive set of performance metrics and nonparametric statistical significance tests; which makes it easier to understand the outcomes and provide a basis for comparisons with related works. In tweets spam modeling, the proposed approach selected less than 10\% of features to attain on average 92.67\% and 82.32\% total accuracy and geometric mean respectively. It outperformed the performance of $Chi^2$ and $PCA$ based approaches in feature selection. In addition, it showed competitive performance compared to recent machine learning algorithms; including word embedding and deep learning based models. 
 
The stochastic aspects of genetic algorithms, and parameter optimization are among the research limitations. Genetic algorithm based solutions usually require a large number of initial population space or large number of generations to find an outperforming solution. The large number of experiment runs and the comprehensive set of performance metrics would direct further research activities. There are many unexplored issues by this research; issues include parallel processing to reduce time complexity of the approach, the effect of natural language processing on improving the accuracy, incorporating user account features in spam modeling, and experimenting with multi-language spam modeling. Further research that may build on the modified genetic algorithm to tackle different problems or domains such as sentiment analysis and multi-class modeling.

\bibliographystyle{IEEEtran}
\bibliography{refs}

\begin{IEEEbiography}[{\includegraphics[width=1in,height=1.25in,clip,keepaspectratio]{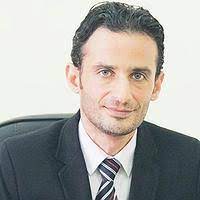}}]{Nazeeh Ghatasheh} is an Associate professor at the Information Technology Department / The University of Jordan (Jordan).  He received his B.Sc. degree in Computer Information Systems from The University of Jordan, Amman, Jordan, in 2004. Then he was awarded merit-based scholarships to pursue his M.Sc. in e-Business Management and Ph.D. in e-Business at the University of Salento (Italy), which he obtained in 2008 and 2011 respectively. He served as Chairman of Information Technology and Computer Information Systems Departments, University Liaison Officer for Quality and Development, Dean’s Assistant for Quality and Development, Faculty Board Member, and Director of the Computer Center at The University of Jordan, Aqaba, Jordan. His research interests include e-Business, Business Analytics, Applied Computational Intelligence, and Data Mining.

\end{IEEEbiography}

\begin{IEEEbiography}[{\includegraphics[width=1in,height=1.25in,clip,keepaspectratio]{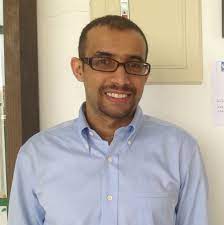}}]{Ismail Altaharwa} 
received his B.Sc. in Computer Science and its applications from the Hashemite University of Jordan in 2005. He continued his studies at AL-Balqaa Applied University, Jordan, earning an M.Sc. in Computer Science (emphasizes in AI techniques especially Computational Intelligence and  Evolutionary Computations) in 2008. He received his PhD in Computer Science and Information Engineering (emphasized in Machine learning techniques and Information Security)  from National Taiwan University of Science and Technology in Taiwan in 2014. Dr. AL-Taharwa is currently an associate professor at the department of computer information systems, The University of Jordan, Aqaba Campus.



\end{IEEEbiography}

\begin{IEEEbiography}[{\includegraphics[width=1in,height=1.25in,clip,keepaspectratio]{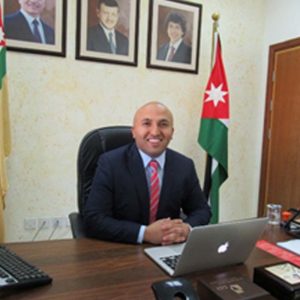}}]{Khaled Aldebei} 
received a B.S. in Software Engineering from Al-Balqa Applied University in 2006, an MS in Computer Science from the same university in 2009, and a Ph.D. in Computer Science from the University of Technology, Sydney, Australia, in 2018. His research interests include natural language processing ($NLP$), machine learning, authorship analysis, data mining and text analysis.



\end{IEEEbiography}

\EOD

\end{document}